\documentclass[cameraready]{template/mac_automl_xmu_blue}
\usepackage{graphicx}
\setleftheadercontent{%
  \headerlogospace{1.6mm}%
  \headerlogo[-0.08\height]{15.2mm}{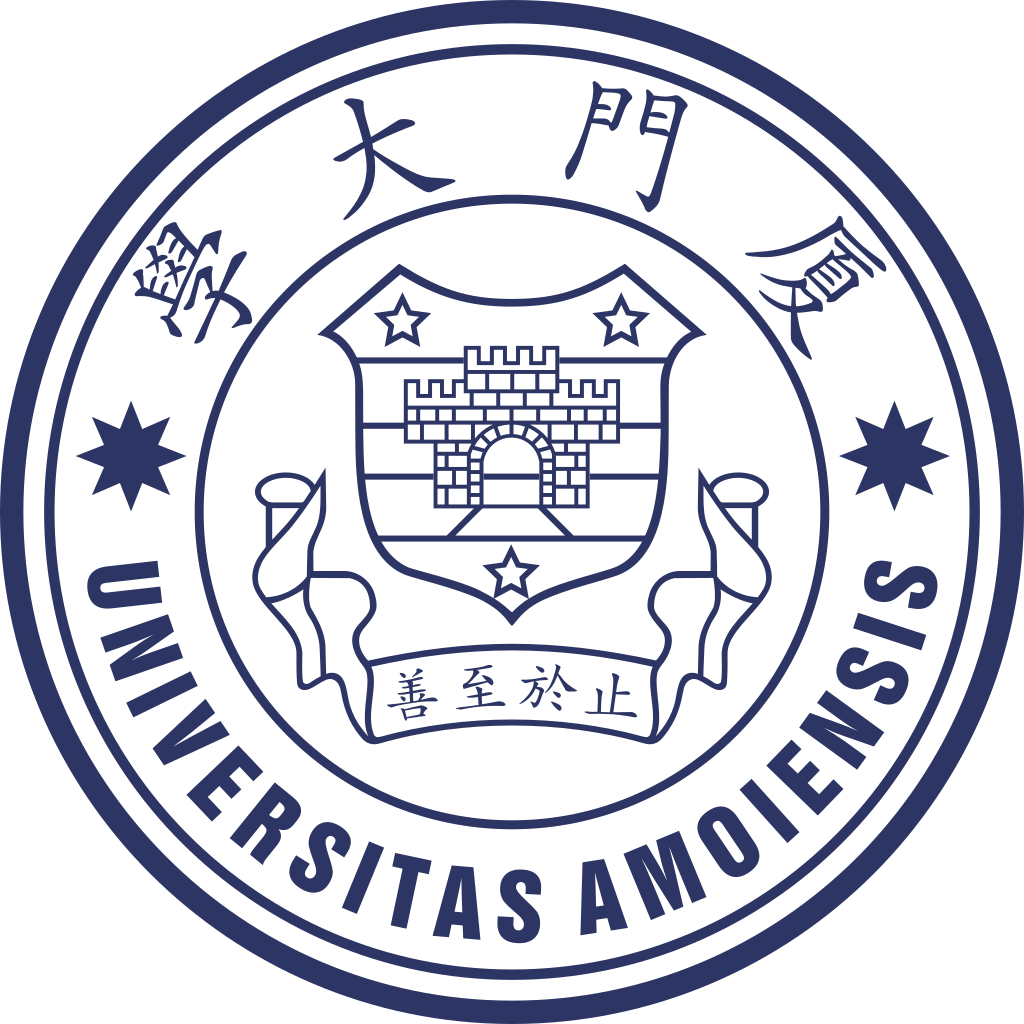}%
  \headerlogospace{2.8mm}%
  \headerlogo[-0.05\height]{12.8mm}{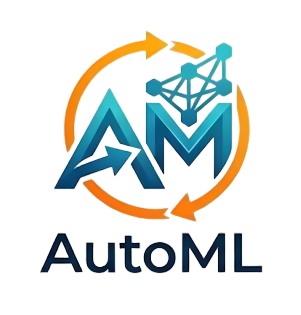}%
}
\setrunningheadericon{%
  \headerlogospace{1.2mm}%
  \headerlogo[-0.03\height]{8.2mm}{assets/branding/automl.png}%
}
\setheadergroupname{MAC-AutoML}
\title{RoTA-Bench: Evaluating 3D Spatial Reasoning through Interactive Projection Alignment}
\setfrontauthors{%
  \authorrow{%
    \authorentry{Qingchuan Ma}{1},\authorsep
    \authorentry{Yuexiao Ma}{2},\authorsep
    \authorentry{Xiang Meng}{1},\authorsep
    \authorentry{Yixuan Zou}{1},\authorsep
    \authorentry{Tianyu Xie}{1}}%
  \authorrow{%
    \authorentry{Wang Chen}{1},\authorsep
    \authorentry{Yuhui Zeng}{1},\authorsep
    \authorentry{Yuhang Wu}{1},\authorsep
    \authorentry{Rongrong Ji}{1,3},\authorsep
    \authorentry{Xiawu Zheng}{1,3,\corremailmark}}
}
\setfrontaffiliations{%
  \affiliationline{\authormark{1}Key Laboratory of Multimedia Trusted Perception and Efficient Computing, Ministry of Education of China, Xiamen University, 361005, P.R. China.}%
  \affiliationline{\authormark{2}s-lab, NTU}%
  \affiliationline{\authormark{3}Institute of Artificial Intelligence, Xiamen University}%
}
\setfrontcontact{%
  \contactline{\textbf{\corremailmark\ Corresponding Author:} \href{mailto:zhengxiawu@xmu.edu.cn}{zhengxiawu@xmu.edu.cn}}%
}
\usecustomauthorlayout
\abstract{Can vision-language models turn 2D observations into 3D understanding that supports precise action? This requires inferring object geometry, anticipating the effects of rotation, and refining actions through visual feedback. We introduce RoTA-Bench to evaluate these abilities through interactive 3D-to-2D projection alignment. Given a target silhouette, a model observes a rotated object and chooses successive rotations to recover the target projection within a limited action budget. Our automated pipeline constructs silhouette-preserving 3D objects from 2D icons and uses reversible initialization to guarantee solvability, yielding 674 tasks and a stratified 100-task test set. Empirically, we find that (i) a substantial human-model gap persists: GPT-5.6-Sol succeeds on only 17\% of tasks, whereas 20 undergraduate participants collectively complete the test set four times with a 100\% success rate; (ii) image-space and pose-space progress can diverge: at matched states, the model more often improves silhouette overlap while worsening pose alignment than humans (19.4\% vs. 10.9\%); and (iii) fine-grained control is sensitive to interaction history: removing it leaves 96.8\% of the model's actions at a coarse 30-degree step size and sharply reduces observed task success. These findings highlight the challenge of converting visual evidence into precise 3D control, beyond recognizing or matching individual images.}
\usepackage{amsmath,amsfonts,bm}

\def\eqref#1{equation~\ref{#1}}
\def\1{\bm{1}}

\DeclareMathAlphabet{\mathsfit}{\encodingdefault}{\sfdefault}{m}{sl}
\SetMathAlphabet{\mathsfit}{bold}{\encodingdefault}{\sfdefault}{bx}{n}

\usepackage{fvextra}
\usepackage{pgfplots}
\usepgfplotslibrary{groupplots,statistics}
\usepackage{pgfplotstable}
\pgfplotsset{compat=1.18}
\definecolor{sapphire}{RGB}{8,115,191}
\definecolor{amber}{RGB}{240,147,43}
\definecolor{primary}{RGB}{0,119,187}
\definecolor{secondary}{RGB}{204,51,17}
\definecolor{tertiary}{RGB}{238,119,51}
\definecolor{quaternary}{RGB}{0,153,136}
\definecolor{research}{RGB}{0,102,51}
\definecolor{lightteal}{RGB}{70,190,150}
\definecolor{goldyellow}{RGB}{253,204,92}
\definecolor{mintgreen}{RGB}{102,217,128}
\begin{document}
\maketitle
\section{Introduction}

Psychological and cognitive-science research establishes spatial perception as a foundational domain of cognitive ability \citep{Carroll_1993}, essential for maintaining object constancy across viewpoints \citep{shepard1971mental} and recovering three-dimensional geometry from two-dimensional projections \citep{richards1987inferring} through the mental representation, manipulation, and inference of shape and orientation. As vision-language models are increasingly applied in visual question answering, interactive agents, and embodied systems \citep{chen2024spatialvlm,wang2026d3d}, agents in interactive settings such as embodied exploration and geometry-aware navigation must translate spatial understanding into physical actions \citep{zhu2025move}. However, current models continue to face challenges in reasoning about three-dimensional positions, orientations, and spatial relationships \citep{ma2025spatialllm}. Therefore, addressing these challenges requires evaluation tools that assess whether models can turn 2D observations into 3D understanding that supports precise action.

However, current visual and spatial benchmarks fall short of this requirement. Static benchmarks evaluate spatial reasoning through passive question answering on fixed images \citep{johnson2017clevr,hudson2019gqa,yue2024mmmu,lu2024mathvista}, lacking action execution and closed-loop feedback entirely. Conversely, existing interactive benchmarks spanning software interfaces, visual mazes, and sliding puzzles \citep{zhou2024webarena,xie2024osworld,rodriguezsalgado2026mazebench,mayer2025ivispar} focus on high-level task planning, tool invocation, or spatial pathfinding in complex environments. In these settings, task success largely depends on coding ability, reasoning, and planning, rather than perceiving three-dimensional orientation from visual cues and translating it into action.

To bridge this gap, inspired by classical insights on inferring three-dimensional shape from two-dimensional projections and silhouettes \citep{richards1987inferring}, we ask whether vision-language models can turn 2D observations into 3D understanding that supports precise action. Therefore, we introduce RoTA-Bench, an automatically constructed benchmark of interactive 3D-to-2D projection alignment. Our pipeline constructs three-dimensional objects from two-dimensional icons, generates initial orientations through reversible rotations, and guarantees solvability automatically. Given a target silhouette and visual observations of the rotated object, the model selects a rotation from 24 discrete actions at each step, receives an updated observation at the next step, and iteratively refines alignment within a limited action budget. The benchmark contains 674 tasks, with a stratified 100-task test set.

Through evaluations of nine vision-language models, two agent configurations, and 20 undergraduate participants, each completing a consecutive 20-task block from the 100-task test set, we find: First, a substantial human--model gap persists: GPT-5.6-Sol succeeds on only 17\% of tasks, whereas all 400 human task runs succeed. Second, image-space and pose-space progress can diverge: at matched states, GPT-5.6-Sol more often improves silhouette overlap while worsening pose alignment than humans (19.4\% vs.\ 10.9\%). Third, fine-grained control is sensitive to interaction history: removing it leaves 96.8\% of the model's actions at a coarse $30^\circ$ step size and sharply reduces observed task success.

\begin{figure*}[t]
\centering
\includegraphics[width=\textwidth]{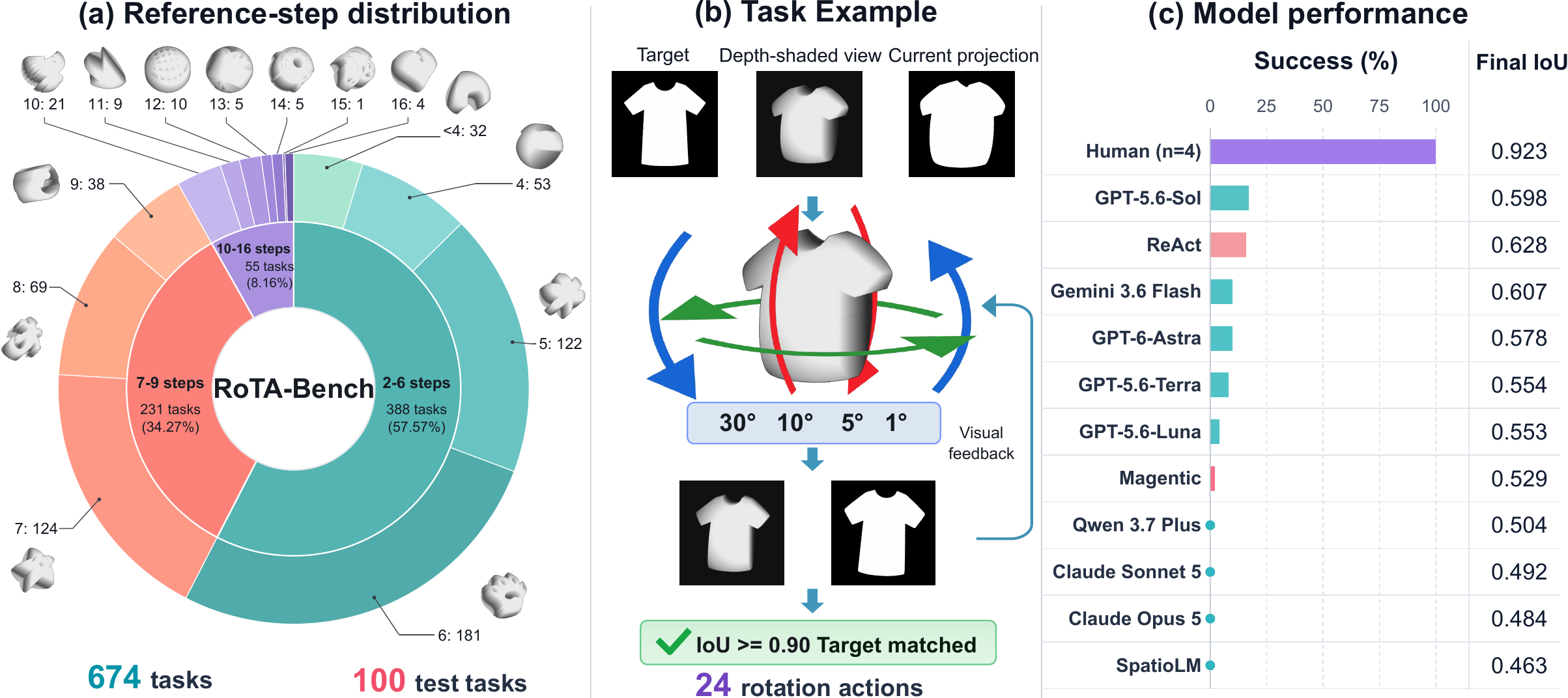}
\caption{RoTA-Bench overview. The benchmark contains 674 canonical tasks, with 100 test tasks selected by stratified sampling over reference path lengths. In the T-shirt example, the solver chooses discrete rotations from visual observations and receives updated views until projection IoU reaches $0.90$ or another stopping rule applies. The right panel shows task success rates and final IoU on the test set; full results are reported in Table~\ref{tab:formal-results}.}
\label{fig:benchmark-overview}
\end{figure*}

\begin{figure*}[t]
\centering
\includegraphics[width=\textwidth]{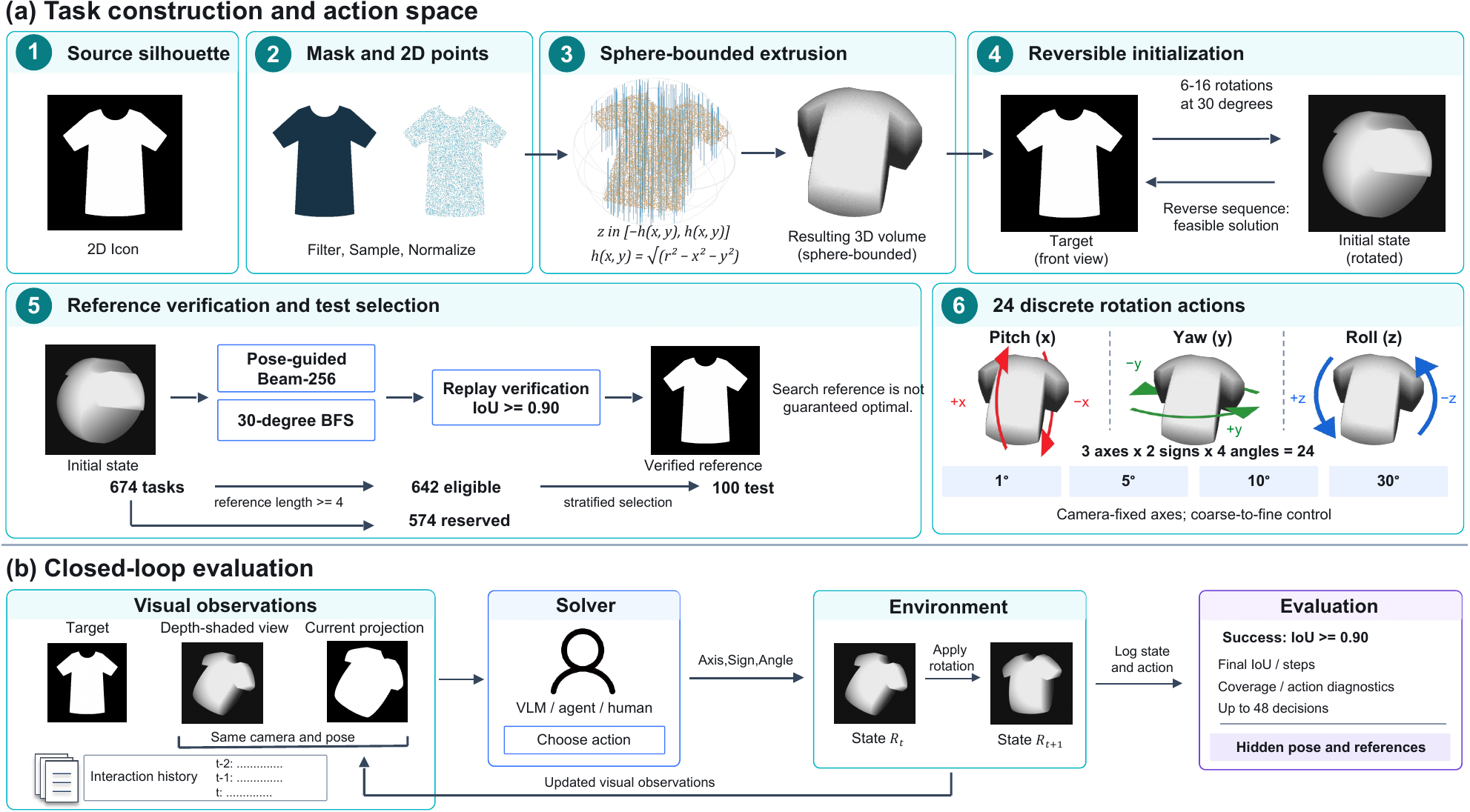}
\caption{RoTA-Bench construction and evaluation pipeline. \textbf{(a)} \textbf{(1)} Source 2D icons are \textbf{(2)} filtered, sampled, and normalized, then \textbf{(3)} converted into 3D objects through sphere-bounded extrusion. \textbf{(4)} Reversible initialization generates an initial pose with a feasible solution. \textbf{(5)} Pose-guided beam search and $30^\circ$-restricted breadth-first search provide verified reference paths, which are used to select 100 test tasks from 674 canonical tasks by stratified sampling over reference path lengths. \textbf{(6)} The action space contains 24 discrete rotations across three axes, two directions, and four angles. \textbf{(b)} In closed-loop evaluation, the solver selects rotations from visual observations and interaction history, then receives updated observations. Success requires projection IoU $\geq 0.90$.}
\label{fig:formal-pipeline}
\end{figure*}

\section{Related Work}

\subsection{Multimodal Large Language Models and Spatial Reasoning}
Early vision-language models established contrastive image--text alignment \citep{radford2021learning}, while subsequent architectures bridged visual encoders with language models for open-ended generation \citep{alayrac2022flamingo,li2023blip} and visual instruction following \citep{liu2023visual}, scaled further by frontier models for advanced visual reasoning \citep{openai2026gpt56systemcard,anthropic2026claudesonnet5,anthropic2026claudeopus5,google2026gemini36flash,qwen37}. To capture 3D geometry, specialized architectures incorporate explicit 3D representations \citep{hong20233d,xu2024pointllm}, whereas spatial VLMs infuse coordinate reference frames \citep{song2025robospatial}, metric relations \citep{chen2024spatialvlm}, 3D object orientations \citep{ma2025spatialllm}, and physical spatial intelligence \citep{wu2026spatiolm}. Embodied and vision-language-action architectures further connect visual reasoning to physical planning \citep{huang2023embodied,rivera2025conceptagent} and robotic execution \citep{driess2023palm,zhao2025cot}. Nonetheless, current multimodal models struggle with camera-pose estimation \citep{deng2026lost}, orthographic projection \citep{niu2026creft}, and shape recognition \citep{eppel2026shape}. Unlike explicit 3D approaches, we examine whether models can turn 2D observations into 3D understanding that supports precise action.

\subsection{Static Benchmarks for Visual and Spatial Reasoning}
Static benchmarks evaluate visual and spatial capabilities on fixed inputs. Foundational suites probe spatial relation queries on synthetic 3D scenes \citep{johnson2017clevr} and natural scene graphs \citep{hudson2019gqa}, alongside multidisciplinary visual diagrams \citep{yue2024mmmu} and geometric problem-solving \citep{lu2024mathvista}. Recent diagnostic benchmarks target 6D object pose \citep{wang2025pulsecheck457} and multi-step action preconditions under complex indoor scenes \citep{xie2026spatialqa}. While A2RBench~\citep{ma2026a2rbench} procedurally generates formally verifiable, multidimensional abstract reasoning tasks, it focuses on symbolic rules rather than grounded 3D spatial perception. Concurrently, offline spatial analyses demonstrate that high performance on static visual mazes often stems from serial search in token space rather than genuine visual planning \citep{rodriguezsalgado2026mazebench}. Crucially, these benchmarks are strictly non-interactive, evaluating spatial reasoning in a single turn without any action or visual feedback.

\subsection{Interactive Benchmarks for Visual and Spatial Reasoning}
Interactive benchmarks evaluate visual and spatial reasoning through multi-turn closed-loop interaction. In embodied AI, interactive benchmarks and agent architectures evaluate spatial reasoning through embodied navigation, long-horizon task execution, and scene exploration \citep{shridhar2020alfred,savva2019habitat,driess2023palm,rana2023sayplan,wang2023voyager,zhu2025move}; however, they predominantly emphasize macroscopic agent locomotion and high-level semantic planning. Across digital domains, benchmarks test 2D layout grounding and interface parsing for web browsing and operating system control \citep{zhou2024webarena,koh2024visualwebarena,xie2024osworld}, yet they operate strictly within 2D planar graphical displays. Closer to geometric reasoning, interactive benchmarks evaluate spatial search via textual coordinates \citep{einarsson2026mazeeval}, open-world 3D labyrinth navigation with Sokoban-style box-pushing puzzles \citep{pappas2026mazebench}, and visual sliding-tile puzzles \citep{mayer2025ivispar}; nevertheless, their objects are axis-aligned grid entities without rotational pose variation, reducing spatial reasoning to combinatorial pathfinding over discrete states. In contrast, RoTA-Bench evaluates whether models can turn 2D observations into 3D understanding that supports precise action through closed-loop projection alignment.

\section{RoTA-Bench}
\label{sec:methods}

RoTA-Bench evaluates whether vision-language models can turn 2D observations into 3D understanding that supports precise action through interactive projection alignment (Figure~\ref{fig:benchmark-overview}). Figure~\ref{fig:formal-pipeline} summarizes task construction and closed-loop evaluation. Below, we detail the benchmark design, task formulation, automated task construction, and evaluation protocol.

\subsection{Motivation and Design Rationale}
\label{sec:methods-motivation}

Classical studies in cognitive vision and psychology demonstrate that underlying 3D geometry and orientations can be inferred from 2D visual contours \citep{richards1987inferring} and compared through mental rotation \citep{shepard1971mental}. Inspired by these foundational insights, we formulate 3D orientation alignment as a projection-matching task: rotating a 3D object until its 2D silhouette matches a target image. Restricting physical interaction to 3D rotation provides two key methodological advantages. First, out-of-plane rotations induce nonlinear silhouette deformations, aspect changes, and self-occlusion, creating an inherently visual challenge that resists textual representation. Second, pure rotation cleanly decouples spatial orientation reasoning from extraneous physical dynamics—such as translational drift, grasp affordances, and contact friction—allowing us to examine whether models can turn 2D observations into 3D understanding that supports precise action.

To reduce reliance on semantic shortcuts and encourage visual-spatial reasoning, RoTA-Bench synthesizes 3D objects from 2D vector icons rather than drawing directly from common 3D CAD repositories, where familiar shapes may trigger pre-trained priors. By extruding 2D icons within a spherical volume, this automated generation produces abstract, novel 3D geometries while mathematically guaranteeing that the canonical front view reproduces the original icon, preserving its fine-grained details (Appendix~\ref{app:target-orientation}). Finally, generating initial poses via sequences of reversible discrete rotations guarantees by construction that every task is solvable.

\subsection{Task Definition}
\label{sec:methods-task-definition}

Following the design rationale in Section~\ref{sec:methods-motivation}, we formulate the interactive projection-matching task as follows. Each task instance is defined by a rigid 3D object $G \subset \mathbb{R}^3$ centered at the coordinate origin, an initial pose $R_0 \in \mathrm{SO}(3)$, and a canonical target pose $R^\star = I_3$. Let $\Pi(\cdot)$ denote the orthographic projection operator that renders the binary silhouette mask of object $G$ at a given pose. At step $t$, the environment provides an observation triplet:
\begin{equation}
O_t = (I^\star,\, V_t,\, I_t), \quad I^\star = \Pi(G, R^\star), \quad I_t = \Pi(G, R_t),
\label{eq:observation-tuple}
\end{equation}
where $I^\star \in \{0, 1\}^{H \times W}$ represents the invariant target silhouette, and $I_t \in \{0, 1\}^{H \times W}$ is the current projection silhouette. To complement the binary silhouette with missing depth and occlusion cues, the agent also receives a depth-shaded auxiliary view $V_t \in [0, 255]^{H \times W}$, rendered from the identical viewpoint as $I_t$.

Conditioned on this visual feedback, the agent interacts with the object through a discrete action space. At each step $t$, the agent selects an action $a_t$ from a vocabulary $\mathcal{A}$ consisting of 24 camera-fixed rotation actions and a terminal STOP command. These 24 actions span six directional senses across three principal camera axes---pitch (up/down), yaw (left/right), and in-plane roll (clockwise/counterclockwise)---paired with four step sizes $\Delta\theta \in \{1^\circ, 5^\circ, 10^\circ, 30^\circ\}$ to enable both coarse and fine pose adjustments. Letting $Q(a_t) \in \mathrm{SO}(3)$ denote the rotation matrix for action $a_t$, the object pose updates via:
\begin{equation}
R_{t+1} = Q(a_t) R_t.
\label{eq:state-transition}
\end{equation}
Executing $a_t$ updates the 3D state and re-renders observations $I_{t+1}$ and $V_{t+1}$.

To evaluate projection alignment, we compute the projection Intersection over Union (projection IoU) between the current projection $I_t$ and the target silhouette $I^\star$:
\begin{equation}
s_t = \operatorname{IoU}(I_t, I^\star) = \frac{|I_t \cap I^\star|}{|I_t \cup I^\star|}.
\label{eq:projection-iou}
\end{equation}
An episode succeeds whenever $s_t \ge \tau$, where we set $\tau = 0.90$. This threshold enforces high visual alignment fidelity while providing a necessary tolerance margin for the discrete action space, where exact alignment is difficult to achieve in finite steps. Furthermore, because geometric symmetries and orthographic self-overlap can produce identical silhouettes from distinct 3D poses, the task naturally admits multiple equivalent target poses. We formalize this goal set as:
\begin{equation}
\mathcal{R}_{\mathrm{goal}}(G, I^\star) = \{ R \in \mathrm{SO}(3) \mid \operatorname{IoU}(\Pi(G, R), I^\star) \ge \tau \}.
\label{eq:goal-set}
\end{equation}
Finally, an episode terminates upon task success ($s_t \ge \tau$), voluntary halting via STOP, reaching a maximum action budget, or stagnation after consecutive steps without improving the best IoU.

\subsection{Dataset Construction}
\label{sec:methods-dataset}

Our automated pipeline constructs the benchmark across three sequential phases: task generation, reference search, and dataset construction, as summarized in Figure~\ref{fig:formal-pipeline}.

First, in the task generation phase, following the design rationale in Section~\ref{sec:methods-motivation}, we curate 2D vector icons from the open-source Game-icons library (Appendix~\ref{app:icon-sources}), rasterizing them into high-resolution binary masks after geometric filtering on connectivity, occupancy, aspect ratio, and foreground point count (Appendix~\ref{app:icon-processing}). To construct a 3D object from each 2D icon while strictly bounding its spatial profile under arbitrary rotations, the planar point set $P \subset \mathbb{R}^2$ is scaled within a disk of radius $r$ and symmetrically extruded along the depth axis under a spherical envelope:
\begin{equation}
h(x, y) = \sqrt{\max(r^2 - x^2 - y^2,\, 0)}, \quad z \in [-h(x, y),\, h(x, y)].
\label{eq:depth-envelope}
\end{equation}
The resulting solid $G = \{(x, y, z) \mid (x, y) \in P,\, |z| \le h(x, y)\}$ is discretized as a dense 3D point cloud (Appendix~\ref{app:icon-processing}). Moreover, because 3D rotations preserve Euclidean norms, any rotated pose satisfies $x^2 + y^2 + z^2 \le r^2$. This spherical invariance prevents viewport clipping and fixes the camera scale globally across all tasks. To provide monocular depth cues and bridge discretization gaps between sampled points, visual observations are rendered via orthographic projection using depth-dependent shading and Gaussian smoothing (Appendix~\ref{app:icon-processing}).

To preserve the original 2D icon's fine contour details (Appendix~\ref{app:target-orientation}), the canonical target pose is fixed to $R^\star = I_3$, and initial poses $R_0 = Q(b_L) \cdots Q(b_1) R^\star$ are generated via $L \sim \text{Uniform}\{6, \dots, 16\}$ discrete rotations. At each step, $b_k \in \mathcal{A}_{30^\circ}$ rotates by $30^\circ$, with its direction sampled uniformly at random from the six candidate camera directions (each with probability $1/6$), forbidding the immediate inverse ($b_k \ne b_{k-1}^{-1}$) to avoid cancellations and guarantee genuine displacement. Initial poses already projection-aligned ($s_0 \ge \tau$) are discarded. Crucially, because every rotation in $\mathrm{SO}(3)$ is invertible, executing the reversed sequence $Q(b_1)^{-1} \cdots Q(b_L)^{-1}$ restores $R^\star$, ensuring 100\% reachability and task solvability by construction. In total, this generation pipeline yields 674 canonical tasks with unique source icons.

Second, in the reference search phase, we compute reference paths using pose-guided beam search and $30^\circ$-restricted breadth-first search. These discover significantly shorter trajectories than raw reversed generation paths (Appendix~\ref{app:search-reference}).

Finally, in the dataset construction phase, a stratified test set of 100 tasks was frozen across reference path lengths ($\ge 4$ steps) using proportional stratified sampling via Hamilton's method and deterministic hash sorting, reserving the remaining 574 tasks for future training and research (Appendix~\ref{app:dataset-construction}).

\subsection{Evaluation Metrics}
\label{sec:methods-metrics}

\paragraph{Task performance metrics.}
Task Success Rate is the fraction of episodes reaching IoU $s_t\geq0.90$. Mean Final IoU and mean rotation steps average the final IoU and executed rotations over all tasks, including failures. An episode fails if it ends through STOP, the 48-decision limit, or stagnation before reaching the threshold.

\paragraph{Geometric diagnostic metrics.}
We compute 3D pose error from ground-truth pose data, which are not provided to the model:
\begin{equation}
e_t = \min_{S \in \mathcal{G}} d_{\mathrm{SO}(3)}(R_t,\, R^\star S) = \min_{S \in \mathcal{G}} \arccos\left(\frac{\operatorname{Tr}(R_t (R^\star S)^\top) - 1}{2}\right).
\label{eq:pose-error}
\end{equation}
Here, $\mathcal{G}$ contains the object's estimated symmetries (Appendix~\ref{app:object-symmetries}). Pose error is the smallest rotation needed to reach an equivalent target pose. Equivalent target poses produce exact alignment, whereas task success requires only a projection IoU of at least $0.90$. Thus, a successful pose need not have zero pose error. Pose error is used as a geometric diagnostic rather than as the task objective; a lower pose error does not necessarily imply a larger immediate IoU improvement. Surface coverage is the fraction of the object's surface observed during an episode.

A direction (axis and sign) is effective if at least one legal angle reduces pose error: $\min_{\theta\in\{1^\circ,5^\circ,10^\circ,30^\circ\}}e(R_t,d,\theta)<e_t$. Effective Direction Rate is the fraction of rotations that choose such a direction; the chosen angle may still increase error. For effective directions, let $\Theta_t^\star=\arg\min_\theta e(R_t,d,\theta)$ be the best legal angles. Best Legal Angle Rate is the fraction choosing one of these angles. Angle Gap is $\min_{\theta\in\Theta_t^\star}|\theta_{\mathrm{selected}}-\theta|$. Angle Loss is the extra pose error caused by the chosen angle: $e(R_t,d,\theta_{\mathrm{selected}})-\min_\theta e(R_t,d,\theta)$.

We also compare changes in IoU and pose error, and correlate rotation angles with both values before each action. Appendix~\ref{app:finding-evidence} gives the definitions and confidence intervals.

\FloatBarrier

% =========================================================================
% 1. 旗舰总表 (按 Final IoU 严格降序排列，彻底解决宽度越界)
% =========================================================================
\begin{table*}[t]
\centering
\small
\setlength{\tabcolsep}{2pt}
\caption{Task alignment performance and action decomposition on 100 test tasks, sorted by Final IoU. Metrics include Task Success Rate (Succ.), final projection IoU, executed steps, visual surface coverage (Cov.), Effective Direction Rate (Eff.\ dir.), and for effective directions: Best Legal Angle Rate (Best ang.), Angle gap, and Angle loss (geometric penalty from sub-optimal step size). SpatioLM abbreviates SpatioLM-Understanding-InternVL3.5.}
\label{tab:formal-results}
\begin{tabular*}{\linewidth}{@{\extracolsep{\fill}} l r r r r r r r r @{}}
\toprule
& \multicolumn{4}{c}{\textbf{Task Alignment Performance}} & \multicolumn{4}{c}{\textbf{Pose Action Decomposition}} \\
\cmidrule(lr){2-5} \cmidrule(lr){6-9}
Method & Succ.\ (\%) & Final IoU & Steps & Cov.\ (\%) & Eff.\ dir.\ (\%) & Best ang.\ (\%) & Gap ($^\circ$) & Loss ($^\circ$) \\
\midrule
Pose greedy     & \textbf{100.0} & \textbf{0.924} & \textbf{8.32} & --    & \textbf{100.0} & \textbf{100.0} & \textbf{0.00} & \textbf{0.00} \\
IoU greedy      & 39.0          & 0.725          & 16.49 & --    & 63.86          & 40.46          & 15.45         & 8.00          \\
\midrule
Human                   & \textbf{100.0} & 0.923          & 16.22 & 83.07 & 83.91          & 71.82          & 2.99          & 0.96          \\
\midrule
ReAct                   & 16.0           & \textbf{0.628} & 24.62 & 79.08 & 59.59          & 58.01          & 7.44          & 2.81          \\
Gemini 3.6 Flash        & 10.0           & 0.607          & 24.68 & 74.72 & 61.87          & 61.69          & 5.47          & 1.99          \\
GPT-5.6-Sol             & \textbf{17.0}  & 0.598          & 23.66 & 79.58 & 58.75          & 61.80          & 6.49          & 2.25          \\
GPT-6-Astra             & 10.0           & 0.578          & 23.18 & 78.93 & 55.78          & 67.36          & 5.85          & 2.14          \\
GPT-5.6-Terra           &  8.0           & 0.554          & 22.91 & 73.13 & 56.31          & 53.80          & 8.50          & 3.58          \\
GPT-5.6-Luna            &  4.0           & 0.553          & 24.29 & 71.56 & 57.64          & 50.86          & 9.15          & 4.03          \\
Random + IoU rollback   &   0.7          & 0.545          & 35.57 & --    &  57.52         &  30.69         & 14.53         & 6.48          \\
Magentic                &  2.0           & 0.529          & 16.00 & 59.71 & 58.31          & 49.95          & 9.77          & 4.12          \\
Qwen 3.7 Plus           &  0.0           & 0.504          & 21.03 & 52.02 & 52.40          & 47.46          & 9.96          & 4.90          \\
Claude Sonnet 5         &  0.0           & 0.492          & 19.39 & 58.66 & 53.38          & 70.53          & 6.10          & 2.15          \\
Claude Opus 5           &  0.0           & 0.484          & 22.09 & 62.40 & 53.28          & 65.68          & 6.85          & 2.29          \\
Uniform random          &   0.0          & 0.473          & 22.23 & --    &  53.78         &  23.45         & 17.05         & 7.67          \\
SpatioLM-InternVL3.5    &  0.0           & 0.463          & 18.81 & 66.85 & 46.15          & 13.36          & 16.55         & 8.61          \\
\bottomrule
\end{tabular*}
\end{table*}

\begin{figure}[t]\centering
\begin{tikzpicture}\begin{axis}[width=\linewidth,height=4.5cm,xmin=0,xmax=48,ymin=0.45,ymax=0.95,xlabel={Rotation steps},ylabel={Mean IoU},grid=major,grid style={gray!15},legend columns=4,legend style={at={(0.5,1)},anchor=south,font=\scriptsize,draw=none},tick label style={font=\normalsize},label style={font=\normalsize}]
\addplot[primary,very thick,mark=*,mark size=0.45pt] coordinates {(0,0.479366) (1,0.490349) (2,0.507435) (3,0.524526) (4,0.550012) (5,0.579473) (6,0.610101) (7,0.641038) (8,0.679650) (9,0.712494) (10,0.747528) (11,0.775764) (12,0.801771) (13,0.821073) (14,0.840848) (15,0.853426) (16,0.867468) (17,0.873713) (18,0.881752) (19,0.885025) (20,0.891388) (21,0.896234) (22,0.898797) (23,0.900753) (24,0.903255) (25,0.905976) (26,0.908164) (27,0.911889) (28,0.912190) (29,0.913024) (30,0.915602) (31,0.916301) (32,0.917991) (33,0.918781) (34,0.919301) (35,0.919809) (36,0.920103) (37,0.920750) (38,0.921431) (39,0.921688) (40,0.922235) (41,0.922485) (42,0.922359) (43,0.922512) (44,0.922706) (45,0.922579) (46,0.922706) (47,0.922829) (48,0.922829)};\addlegendentry{Human}
\addplot[research,thick,mark=*,mark size=0.45pt] coordinates {(0,0.479366) (1,0.487406) (2,0.499194) (3,0.494138) (4,0.503406) (5,0.498089) (6,0.507787) (7,0.521756) (8,0.524483) (9,0.531760) (10,0.535245) (11,0.541298) (12,0.549964) (13,0.567797) (14,0.568123) (15,0.571872) (16,0.578353) (17,0.579960) (18,0.585781) (19,0.591775) (20,0.601767) (21,0.602368) (22,0.604447) (23,0.607357) (24,0.609909) (25,0.611763) (26,0.611194) (27,0.614754) (28,0.615326) (29,0.614077) (30,0.618858) (31,0.619864) (32,0.618470) (33,0.623932) (34,0.624127) (35,0.623879) (36,0.624740) (37,0.626270) (38,0.626294) (39,0.626656) (40,0.625190) (41,0.626208) (42,0.626482) (43,0.627863) (44,0.628128) (45,0.627409) (46,0.626716) (47,0.627297) (48,0.627535)};\addlegendentry{ReAct}
\addplot[quaternary,thick,mark=*,mark size=0.45pt] coordinates {(0,0.479366) (1,0.486406) (2,0.497313) (3,0.511595) (4,0.507398) (5,0.517500) (6,0.526047) (7,0.533953) (8,0.542366) (9,0.547178) (10,0.551841) (11,0.552228) (12,0.556986) (13,0.561170) (14,0.567525) (15,0.567826) (16,0.568689) (17,0.577880) (18,0.578634) (19,0.587596) (20,0.584103) (21,0.595598) (22,0.601293) (23,0.602194) (24,0.593833) (25,0.593718) (26,0.594378) (27,0.591539) (28,0.596140) (29,0.603546) (30,0.603484) (31,0.603474) (32,0.603542) (33,0.608061) (34,0.606794) (35,0.607879) (36,0.609715) (37,0.607706) (38,0.605654) (39,0.610506) (40,0.611922) (41,0.611228) (42,0.608773) (43,0.610338) (44,0.609937) (45,0.610157) (46,0.607834) (47,0.607699) (48,0.606636)};\addlegendentry{Gemini 3.6 Flash}
\addplot[secondary,very thick,mark=*,mark size=0.45pt] coordinates {(0,0.479366) (1,0.485870) (2,0.493117) (3,0.496164) (4,0.493673) (5,0.501159) (6,0.506994) (7,0.510593) (8,0.523287) (9,0.527934) (10,0.527470) (11,0.530694) (12,0.538298) (13,0.536679) (14,0.546684) (15,0.542583) (16,0.545485) (17,0.550569) (18,0.560220) (19,0.564595) (20,0.570617) (21,0.569933) (22,0.572639) (23,0.579883) (24,0.576015) (25,0.582004) (26,0.588077) (27,0.585829) (28,0.587820) (29,0.587756) (30,0.587395) (31,0.590373) (32,0.590558) (33,0.589467) (34,0.591491) (35,0.592558) (36,0.594656) (37,0.594136) (38,0.595032) (39,0.593848) (40,0.596844) (41,0.596541) (42,0.598220) (43,0.597870) (44,0.597508) (45,0.596100) (46,0.597357) (47,0.597764) (48,0.598438)};\addlegendentry{GPT-5.6-Sol}
\addplot[amber,thick,mark=*,mark size=0.45pt] coordinates {(0,0.479366) (1,0.485303) (2,0.494303) (3,0.492854) (4,0.493522) (5,0.499403) (6,0.504869) (7,0.503038) (8,0.500259) (9,0.502064) (10,0.504872) (11,0.506006) (12,0.514032) (13,0.520504) (14,0.524025) (15,0.525513) (16,0.528861) (17,0.537687) (18,0.540698) (19,0.541636) (20,0.547300) (21,0.543268) (22,0.546642) (23,0.545819) (24,0.547768) (25,0.549287) (26,0.549819) (27,0.548142) (28,0.549751) (29,0.550850) (30,0.549534) (31,0.549791) (32,0.548980) (33,0.549257) (34,0.551186) (35,0.551303) (36,0.551055) (37,0.550090) (38,0.549655) (39,0.548786) (40,0.548473) (41,0.548652) (42,0.549897) (43,0.551225) (44,0.551446) (45,0.551834) (46,0.553219) (47,0.553481) (48,0.554430)};\addlegendentry{GPT-5.6-Terra}
\addplot[tertiary,thick,mark=*,mark size=0.45pt] coordinates {(0,0.479366) (1,0.485764) (2,0.488430) (3,0.491355) (4,0.489354) (5,0.495343) (6,0.503650) (7,0.507429) (8,0.512644) (9,0.513710) (10,0.512968) (11,0.513666) (12,0.513980) (13,0.519901) (14,0.524290) (15,0.520308) (16,0.522158) (17,0.530616) (18,0.537667) (19,0.542340) (20,0.540214) (21,0.543295) (22,0.545026) (23,0.546468) (24,0.547656) (25,0.549610) (26,0.551569) (27,0.551808) (28,0.550767) (29,0.549954) (30,0.549843) (31,0.550626) (32,0.553481) (33,0.554377) (34,0.554934) (35,0.553849) (36,0.553410) (37,0.553186) (38,0.554366) (39,0.552368) (40,0.553777) (41,0.554431) (42,0.554704) (43,0.553746) (44,0.555159) (45,0.555380) (46,0.554788) (47,0.553324) (48,0.553018)};\addlegendentry{GPT-5.6-Luna}
\addplot[mintgreen,thick,mark=*,mark size=0.45pt] coordinates {(0,0.479366) (1,0.485828) (2,0.487760) (3,0.491691) (4,0.493378) (5,0.496257) (6,0.496390) (7,0.502705) (8,0.506342) (9,0.503408) (10,0.505958) (11,0.506331) (12,0.509732) (13,0.512488) (14,0.516330) (15,0.517640) (16,0.516240) (17,0.516487) (18,0.518191) (19,0.520953) (20,0.521616) (21,0.525094) (22,0.525766) (23,0.529432) (24,0.530057) (25,0.529825) (26,0.530717) (27,0.529717) (28,0.530683) (29,0.528022) (30,0.527251) (31,0.532721) (32,0.532865) (33,0.532739) (34,0.531951) (35,0.533243) (36,0.533156) (37,0.532895) (38,0.533035) (39,0.531617) (40,0.531533) (41,0.530371) (42,0.530681) (43,0.529451) (44,0.529431) (45,0.528201) (46,0.526917) (47,0.528062) (48,0.529265)};\addlegendentry{Magentic}
\addplot[goldyellow,thick,mark=*,mark size=0.45pt] coordinates {(0,0.479366) (1,0.479832) (2,0.484516) (3,0.485626) (4,0.487103) (5,0.490067) (6,0.497189) (7,0.502825) (8,0.499757) (9,0.504176) (10,0.503787) (11,0.506856) (12,0.506092) (13,0.511330) (14,0.508987) (15,0.509531) (16,0.506764) (17,0.505658) (18,0.508560) (19,0.508813) (20,0.508246) (21,0.505636) (22,0.505985) (23,0.504701) (24,0.503173) (25,0.504966) (26,0.503952) (27,0.504522) (28,0.505641) (29,0.504357) (30,0.506657) (31,0.505118) (32,0.505547) (33,0.505154) (34,0.504586) (35,0.505294) (36,0.505044) (37,0.504537) (38,0.504533) (39,0.504514) (40,0.504575) (41,0.504637) (42,0.504694) (43,0.504509) (44,0.504278) (45,0.504096) (46,0.503914) (47,0.503735) (48,0.503735)};\addlegendentry{Qwen 3.7 Plus}
\addplot[secondary,thick,mark=*,mark size=0.45pt] coordinates {(0,0.479366) (1,0.473070) (2,0.479417) (3,0.483408) (4,0.490509) (5,0.492744) (6,0.487399) (7,0.492837) (8,0.489406) (9,0.485842) (10,0.491889) (11,0.483155) (12,0.486479) (13,0.487047) (14,0.484245) (15,0.493462) (16,0.489220) (17,0.489515) (18,0.493462) (19,0.490600) (20,0.489476) (21,0.488197) (22,0.495325) (23,0.493925) (24,0.494870) (25,0.492416) (26,0.492253) (27,0.490373) (28,0.489319) (29,0.490875) (30,0.490224) (31,0.491843) (32,0.490741) (33,0.491885) (34,0.491959) (35,0.490235) (36,0.490995) (37,0.491774) (38,0.491185) (39,0.491857) (40,0.491446) (41,0.490831) (42,0.490605) (43,0.491249) (44,0.491652) (45,0.490866) (46,0.491500) (47,0.491500) (48,0.491500)};\addlegendentry{Claude Sonnet 5}
\addplot[lightteal,thick,mark=*,mark size=0.45pt] coordinates {(0,0.479366) (1,0.475450) (2,0.480767) (3,0.483341) (4,0.483686) (5,0.483311) (6,0.481355) (7,0.483565) (8,0.482284) (9,0.481002) (10,0.485702) (11,0.487240) (12,0.488148) (13,0.488117) (14,0.489681) (15,0.489446) (16,0.485822) (17,0.488793) (18,0.487978) (19,0.487494) (20,0.486412) (21,0.485265) (22,0.486645) (23,0.487169) (24,0.485566) (25,0.484426) (26,0.482107) (27,0.482913) (28,0.483854) (29,0.483799) (30,0.481762) (31,0.482802) (32,0.481223) (33,0.480929) (34,0.481179) (35,0.482139) (36,0.482389) (37,0.483086) (38,0.483431) (39,0.483429) (40,0.483467) (41,0.483096) (42,0.483096) (43,0.483115) (44,0.483083) (45,0.483111) (46,0.483057) (47,0.483321) (48,0.483828)};\addlegendentry{Claude Opus 5}
\addplot[sapphire,thin,dotted,mark=*,mark size=0.45pt] coordinates {(0,0.479366) (1,0.479614) (2,0.479242) (3,0.478007) (4,0.477945) (5,0.477690) (6,0.477844) (7,0.478391) (8,0.478482) (9,0.478807) (10,0.479388) (11,0.480984) (12,0.481462) (13,0.480328) (14,0.479830) (15,0.479730) (16,0.479425) (17,0.478839) (18,0.477842) (19,0.476796) (20,0.477772) (21,0.476962) (22,0.476341) (23,0.475395) (24,0.475499) (25,0.473932) (26,0.474511) (27,0.474262) (28,0.474070) (29,0.473378) (30,0.473631) (31,0.473784) (32,0.473558) (33,0.473410) (34,0.473181) (35,0.473230) (36,0.473662) (37,0.473717) (38,0.474040) (39,0.473588) (40,0.473556) (41,0.473156) (42,0.472683) (43,0.472626) (44,0.472542) (45,0.472484) (46,0.472368) (47,0.472563) (48,0.472819)};\addlegendentry{Uniform random}
\addplot[amber,thin,dashdotted,mark=*,mark size=0.45pt] coordinates {(0,0.479361) (1,0.479612) (2,0.484160) (3,0.484300) (4,0.487412) (5,0.488874) (6,0.492254) (7,0.494636) (8,0.495340) (9,0.497670) (10,0.498386) (11,0.499432) (12,0.502203) (13,0.503624) (14,0.504042) (15,0.506957) (16,0.509157) (17,0.510289) (18,0.509610) (19,0.513721) (20,0.515556) (21,0.514204) (22,0.517305) (23,0.518464) (24,0.520797) (25,0.522893) (26,0.525738) (27,0.524497) (28,0.526951) (29,0.527916) (30,0.529808) (31,0.530222) (32,0.532376) (33,0.529769) (34,0.534576) (35,0.534067) (36,0.536229) (37,0.536061) (38,0.537340) (39,0.537672) (40,0.539405) (41,0.540018) (42,0.541171) (43,0.541010) (44,0.541096) (45,0.544186) (46,0.543475) (47,0.547455) (48,0.544647)};\addlegendentry{Random + IoU rollback}
\addplot[black!60,thin,mark=*,mark size=0.45pt] coordinates {(0,0.479366) (1,0.517242) (2,0.555757) (3,0.593653) (4,0.633421) (5,0.664390) (6,0.684911) (7,0.700416) (8,0.709451) (9,0.715417) (10,0.718463) (11,0.719639) (12,0.723334) (13,0.723017) (14,0.723435) (15,0.723086) (16,0.723479) (17,0.723137) (18,0.723580) (19,0.723332) (20,0.723774) (21,0.723626) (22,0.723923) (23,0.723867) (24,0.724266) (25,0.724422) (26,0.724560) (27,0.724502) (28,0.724632) (29,0.724537) (30,0.724663) (31,0.724686) (32,0.724683) (33,0.724703) (34,0.724707) (35,0.724704) (36,0.724710) (37,0.724709) (38,0.724710) (39,0.724709) (40,0.724710) (41,0.724709) (42,0.724710) (43,0.724709) (44,0.724710) (45,0.724709) (46,0.724710) (47,0.724712) (48,0.724710)};\addlegendentry{IoU oracle}
\addplot[black,thick,dashdotted,mark=*,mark size=0.45pt] coordinates {(0,0.479366) (1,0.493864) (2,0.522889) (3,0.573808) (4,0.635672) (5,0.703664) (6,0.773937) (7,0.832801) (8,0.870725) (9,0.896253) (10,0.909456) (11,0.915755) (12,0.920068) (13,0.922191) (14,0.923562) (15,0.923767) (16,0.924140) (17,0.924140) (18,0.924140) (19,0.924140) (20,0.924140) (21,0.924140) (22,0.924140) (23,0.924140) (24,0.924140) (25,0.924140) (26,0.924140) (27,0.924140) (28,0.924140) (29,0.924140) (30,0.924140) (31,0.924140) (32,0.924140) (33,0.924140) (34,0.924140) (35,0.924140) (36,0.924140) (37,0.924140) (38,0.924140) (39,0.924140) (40,0.924140) (41,0.924140) (42,0.924140) (43,0.924140) (44,0.924140) (45,0.924140) (46,0.924140) (47,0.924140) (48,0.924140)};\addlegendentry{Pose oracle}
\addplot[orange!75!black,thick,mark=*,mark size=0.45pt] coordinates {(0,0.479366) (1,0.481782) (2,0.488223) (3,0.490324) (4,0.494609) (5,0.491518) (6,0.489432) (7,0.496831) (8,0.498279) (9,0.498776) (10,0.498149) (11,0.506328) (12,0.518654) (13,0.522709) (14,0.530170) (15,0.531004) (16,0.540534) (17,0.547465) (18,0.546407) (19,0.547447) (20,0.548545) (21,0.549490) (22,0.554792) (23,0.556756) (24,0.558644) (25,0.564865) (26,0.563984) (27,0.561623) (28,0.562924) (29,0.564750) (30,0.564551) (31,0.566665) (32,0.567740) (33,0.567479) (34,0.574664) (35,0.576006) (36,0.575077) (37,0.575528) (38,0.575605) (39,0.578259) (40,0.579939) (41,0.578867) (42,0.577864) (43,0.577427) (44,0.574975) (45,0.578437) (46,0.579807) (47,0.578263) (48,0.578272)};\addlegendentry{GPT-6-Astra}
\addplot[teal!70!black,thick,mark=*,mark size=0.45pt] coordinates {(0,0.479366) (1,0.480222) (2,0.480459) (3,0.481764) (4,0.480677) (5,0.477978) (6,0.478580) (7,0.477942) (8,0.475724) (9,0.475016) (10,0.473000) (11,0.470230) (12,0.472117) (13,0.472542) (14,0.472845) (15,0.471543) (16,0.466896) (17,0.463293) (18,0.462829) (19,0.462747) (20,0.462748) (21,0.463965) (22,0.464087) (23,0.464564) (24,0.464974) (25,0.465476) (26,0.465322) (27,0.465290) (28,0.465149) (29,0.464970) (30,0.464804) (31,0.464393) (32,0.464047) (33,0.463724) (34,0.463483) (35,0.463310) (36,0.463163) (37,0.463063) (38,0.463033) (39,0.462973) (40,0.462955) (41,0.462953) (42,0.462955) (43,0.462955) (44,0.462955) (45,0.462955) (46,0.462955) (47,0.462955) (48,0.462955)};\addlegendentry{SpatioLM-Understanding-InternVL3.5}
\addplot[gray,densely dashed,forget plot] coordinates {(0,.9) (48,.9)};
\end{axis}\end{tikzpicture}
\caption{Mean IoU versus executed rotations for all methods with exported step trajectories on the same 100 tasks, including the 400 Human task runs and both extension runs. Each marker is the mean at one integer step (0--48), joined without smoothing; terminal IoU is retained after termination through step 48. The dashed line marks the success threshold.}
\label{fig:mean-iou-main}
\end{figure}

\section{Experimental Setup}
\label{sec:experiments}

We evaluate nine vision-language models---GPT-5.6-Sol, GPT-5.6-Terra, GPT-5.6-Luna \citep{openai2026gpt56systemcard}, Gemini 3.6 Flash \citep{google2026gemini36flash}, Claude Opus 5, Claude Sonnet 5 \citep{anthropic2026claudeopus5,anthropic2026claudesonnet5}, Qwen 3.7 Plus \citep{qwen37}, GPT-6-Astra \citep{openai2026gpt6systemcard}, and SpatioLM-Understanding-InternVL3.5 \citep{wu2026spatiolm}---as well as two GPT-5.6-Sol-based agent frameworks, ReAct \citep{yao2022react} and Magentic \citep{fourney2024magentic}, 20 undergraduate participants (Human; Appendix~\ref{app:human-evaluation}), two stochastic baselines (Uniform random and Random with IoU rollback), and two greedy baselines. All evaluations follow a common task protocol, action space, decision budget, success criterion, and termination rules. All models are evaluated at temperature 0; closed-source models are queried via APIs (total cost: \$1,482.07), while SpatioLM is deployed locally on a server equipped with eight NVIDIA A100 (40GB) GPUs. Complete prompts, hyperparameters, agent tools, and evaluation settings are detailed in Appendices~\ref{app:implementation} and~\ref{app:prompts}.

\section{Results}
\label{sec:results}
In this section, we present main results, conduct rotation direction and angle analysis, perform an ablation of auxiliary views and interaction history, analyze action association with projection and pose discrepancies, and discuss limitations and future work.

\subsection{Main Results}
\label{sec:results-main}

Table~\ref{tab:formal-results} presents alignment performance across 100 test tasks, with success counts and confidence intervals in Appendix~\ref{app:success-intervals}. Frontier models struggle markedly on interactive 3D projection alignment: GPT-5.6-Sol achieves the highest success rate at only 17.0\%, followed by Gemini 3.6 Flash at 10.0\%, while ReAct attains the highest final IoU of 0.628 (16.0\% success). Claude Sonnet 5, Claude Opus 5, and Qwen 3.7 Plus fail entirely (0.0\% success), with final IoU values (0.484--0.504) barely exceeding uniform random exploration (0.473). Notably, SpatioLM-Understanding-InternVL3.5 suffers a catastrophic policy collapse, where 91.12\% of its actions degenerate into \texttt{YAW\_POS 10$^\circ$}, yielding 0.0\% success and the lowest final IoU of 0.463 (below random actions). In stark contrast, human participants achieve a 100.0\% success rate, reaching an average final IoU of 0.923 in 16.22 steps. As shown in Figure~\ref{fig:mean-iou-main}, human trajectories exhibit overall IoU gains surpassing the 0.90 threshold, whereas model trajectories plateau early between 0.50 and 0.63. The two greedy baselines sharpen this distinction: at each step, each enumerates all 24 legal rotation successors using the ground-truth simulator, then selects the action that maximizes successor projection IoU or minimizes true 3D pose error, respectively. The IoU-greedy baseline succeeds on only 39.0\% of tasks (final IoU 0.725), showing that greedily maximizing immediate projection overlap is insufficient on many tasks; by contrast, the pose-greedy baseline solves every task (100.0\%, final IoU 0.924 in 8.32 steps), showing that reliable access to the true 3D pose error is sufficient to complete all of them. Overall, the success rates and IoU trajectories show a substantial human--model gap on RoTA-Bench.

\subsection{Diagnosing Model Failures}
\label{sec:results-failures}

We first evaluate whether model failure stems from inadequate visual exploration. As reported in Table~\ref{tab:formal-results}, GPT-5.6-Sol achieves a cumulative surface coverage of 79.58\%, closely trailing human participants (83.07\%), while ReAct (79.08\%) and GPT-6-Astra (78.93\%) maintain comparable exposure. Total coverage is similar, but the regions viewed and their use may differ.

To diagnose model action failures, we decouple each rotation into direction (rotation axis and sign) and angle (step size), which correspond to estimating the direction and magnitude of the 3D pose difference between the current pose and target pose. Evaluating these decomposed components in Table~\ref{tab:formal-results} reveals clear deficiencies on both fronts:
For directional choice, GPT-5.6-Sol selects effective rotation directions far less often than humans (58.75\% vs.\ 83.91\%) and performs only marginally better than uniform random sampling (53.78\%), with other frontier models exhibiting similarly low rates (52.40\%--61.87\%). 
For angular precision, conditioned on choosing an effective direction, GPT-5.6-Sol exhibits a larger mean deviation between its selected angle and the best legal angle (6.49$^\circ$ vs.\ 2.99$^\circ$ for humans; best legal angle rate 61.80\% vs.\ 71.82\%), incurring an average geometric angle loss of 2.25$^\circ$ per step (more than double the 0.96$^\circ$ loss of humans). 
Appendix~\ref{app:finding-one} reports paired confidence intervals for the Human--Sol differences in Effective Direction Rate and Angle Gap. Together, these action metrics show that the human--model gap also appears in rotation direction and angle selection.

% =========================================================================
% 2. 输入消融实验表 (按 Final IoU 降序排列)
% =========================================================================
\begin{table*}[t]
\centering
\small
\setlength{\tabcolsep}{3.5pt}
\caption{Input ablation for GPT-5.6-Sol on 30 test tasks, with ablation conditions sorted by Final IoU. Columns report Task Success Rate (Succ.), final projection IoU, Effective Direction Rate (Eff.\ dir.), percentage distribution across the four legal rotation angles, and Spearman correlation ($\rho$) between pre-action IoU and rotation angle magnitude with 95\% confidence intervals.}
\label{tab:input_ablation}
\begin{tabular*}{\linewidth}{@{\extracolsep{\fill}}lccccccccc@{}}
\toprule
& Success & Final & Effective & \multicolumn{4}{c}{Rotation angle (\%)} & \multicolumn{2}{c}{IoU--angle correlation} \\
\cmidrule(lr){5-8} \cmidrule(lr){9-10}
Condition & (\%) & IoU & direction (\%) & $1^\circ$ & $5^\circ$ & $10^\circ$ & $30^\circ$ & $\rho$ & 95\% CI \\
\midrule
A: Full input        & \textbf{16.67} & \textbf{0.5656} & \textbf{56.52} & 12.18 & 7.51 & 17.56 & 62.75 & -0.472 & [-0.626, -0.272] \\
B: No aux. view &  3.33          & 0.5490          & 55.90          &  3.54 & 3.24 &  7.08 & 86.14 & -0.463 & [-0.605, -0.238] \\
D: Neither input     &  6.67          & 0.5087          & 55.02          &  1.29 & 1.29 &  0.81 & 96.60 & -0.295 & [-0.468, -0.066] \\
C: No history        &  0.00          & 0.4822          & 54.85          &  0.16 & 0.49 &  2.59 & 96.76 & -0.242 & [-0.331, -0.131] \\
\midrule
Uniform random       &  0.00          & 0.4808          & 53.12          & 24.24 & 25.27 & 27.10 & 23.39 & -0.027 & [-0.073,  0.019] \\
\bottomrule
\end{tabular*}
\end{table*}

\subsection{Ablation: Auxiliary Views and Interaction History}
\label{sec:results-ablation}

We evaluate the contribution of visual depth cues and multi-turn feedback by ablating the auxiliary depth view and interaction history for GPT-5.6-Sol across the first 30 tasks, whose distribution of reference solution steps strongly aligns with the 100-task test set (Spearman's $\rho = 0.9121$, $p = 3.56 \times 10^{-5}$; 10,000 bootstrap resamples yield a 95\% CI lower bound of $0.7445$). As reported in Table~\ref{tab:input_ablation}, removing the auxiliary view reduces task success from 16.67\% (full input) to 3.33\% and final IoU from 0.5656 to 0.5490. Removing interaction history reduces observed success from 5/30 tasks to 0/30 and final IoU from 0.5656 to 0.4822, close to uniform random performance (0.4808). The observed effect of removing history is larger than that of removing the auxiliary view.

Interestingly, across ablation settings, the effective-direction rate remains nearly unchanged (54.85\%--56.52\%), while angle selection changes markedly. Removing the auxiliary depth view increases $30^\circ$ rotations to 86.14\% but leaves the IoU--angle correlation close to that under full input ($\rho=-0.463$ vs.\ $-0.472$). Without interaction history, $30^\circ$ rotations rise to 96.76\% from 62.75\%, $1^\circ$ adjustments fall to 0.16\% from 12.18\%, and the correlation weakens to $\rho=-0.242$. In this ablation, history helps the model choose how far to rotate more than which way to rotate. Previous views and actions may help it recognize when to follow a coarse turn with a finer adjustment. Counts and paired intervals are in Appendix~\ref{app:finding-two}. These results reveal that GPT-5.6-Sol's fine-grained control is sensitive to interaction history.

\begin{figure*}[ht]
\centering
\begin{tikzpicture}[x=1.05cm,y=2.5cm]
\foreach \v in {0,0.2,0.4,0.6,0.8} {
    \draw[gray!20] (-0.55,\v) -- (11.55,\v);
    \node[font=\scriptsize,anchor=east] at (-0.65,\v) {\v};
}
\draw[gray!70] (-0.55,0) -- (11.55,0);
\node[font=\small,rotate=90] at (-1.35,0.4) {$|\rho|$};

\fill[sapphire!80] (3.5,0.87) rectangle (3.75,0.90);
\node[font=\small,anchor=west] at (3.85,0.885) {IoU--angle};
\fill[amber!85] (6.6,0.87) rectangle (6.85,0.90);
\node[font=\small,anchor=west] at (6.95,0.885) {Pose--angle};

% name / x / |IoU rho| / CI low / CI high / |Pose rho| / CI low / CI high
\foreach \name/\x/\i/\ilo/\ihi/\p/\plo/\phi in {
    {Human}/0/0.696/0.633/0.763/0.751/0.675/0.828,
    {ReAct}/1/0.570/0.453/0.674/0.320/0.108/0.512,
    {Gemini 3.6 Flash}/2/0.562/0.480/0.633/0.618/0.496/0.709,
    {GPT-5.6-Sol}/3/0.536/0.443/0.617/0.368/0.166/0.532,
    {GPT-6-Astra}/4/0.533/0.427/0.621/0.452/0.266/0.607,
    {GPT-5.6-Terra}/5/0.438/0.335/0.526/0.254/0.056/0.440,
    {GPT-5.6-Luna}/6/0.471/0.373/0.561/0.280/0.088/0.470,
    {Magentic}/7/0.368/0.215/0.493/0.250/0.066/0.403,
    {Qwen 3.7 Plus}/8/0.383/0.251/0.495/0.380/0.232/0.514,
    {Claude Sonnet 5}/9/0.255/0.140/0.365/0.154/0/0.309,
    {Claude Opus 5}/10/0.091/0/0.229/0.204/0.043/0.354,
    {SpatioLM}/11/0.127/0/0.309/0.142/0/0.268
} {
    \fill[sapphire!80] ({\x-0.29},0) rectangle ({\x-0.04},\i);
    \fill[amber!85] ({\x+0.04},0) rectangle ({\x+0.29},\p);

    \draw[black,line width=0.55pt] ({\x-0.165},\ilo) -- ({\x-0.165},\ihi);
    \draw[black,line width=0.55pt] ({\x-0.24},\ilo) -- ({\x-0.09},\ilo);
    \draw[black,line width=0.55pt] ({\x-0.24},\ihi) -- ({\x-0.09},\ihi);

    \draw[black,line width=0.55pt] ({\x+0.165},\plo) -- ({\x+0.165},\phi);
    \draw[black,line width=0.55pt] ({\x+0.09},\plo) -- ({\x+0.24},\plo);
    \draw[black,line width=0.55pt] ({\x+0.09},\phi) -- ({\x+0.24},\phi);

    \node[font=\scriptsize,rotate=25,anchor=north east] at (\x,0.055) {\name};
}
\end{tikzpicture}
\caption{Absolute Spearman correlations between selected rotation angles and pre-action projection IoU or 3D pose error across the 100 test tasks. Bars show $|\rho|$; whiskers show the original 95\% bootstrap intervals mapped to absolute values. An interval crossing zero has a lower bound of zero after this transformation. SpatioLM abbreviates SpatioLM-Understanding-InternVL3.5.}
\label{fig:all-method-correlations}
\end{figure*}

\begin{figure*}[ht]
\centering
\begin{minipage}[c]{0.53\linewidth}
\centering
\begin{tikzpicture}[x=0.058cm,y=0.8cm]
% Light guide lines
\foreach \x in {0,25,50,75,100} {
    \draw[gray!20] (\x,0.35) -- (\x,2.45);
    \node[font=\scriptsize,anchor=north] at (\x,0.25) {\x};
}
\node[font=\scriptsize,anchor=east] at (-3,1.95) {Human};
\node[font=\scriptsize,anchor=east] at (-3,0.95) {GPT-5.6-Sol};

% Human: UC 50.24, DC 33.18, UF 10.90, DF 5.69
\fill[sapphire!85] (0,1.72) rectangle (50.24,2.18);
\fill[lightteal!85] (50.24,1.72) rectangle (83.42,2.18);
\fill[amber!85] (83.42,1.72) rectangle (94.32,2.18);
\fill[secondary!80] (94.32,1.72) rectangle (100,2.18);

% GPT-5.6-Sol: UC 34.60, DC 23.70, UF 19.43, DF 22.27
\fill[sapphire!85] (0,0.72) rectangle (34.60,1.18);
\fill[lightteal!85] (34.60,0.72) rectangle (58.30,1.18);
\fill[amber!85] (58.30,0.72) rectangle (77.73,1.18);
\fill[secondary!80] (77.73,0.72) rectangle (100,1.18);

% Compact legend
\fill[sapphire!85] (2,-0.55) rectangle (7,-0.31);
\node[font=\scriptsize,anchor=west] at (9,-0.43) {UC};
\fill[lightteal!85] (24,-0.55) rectangle (29,-0.31);
\node[font=\scriptsize,anchor=west] at (31,-0.43) {DC};
\fill[amber!85] (46,-0.55) rectangle (51,-0.31);
\node[font=\scriptsize,anchor=west] at (53,-0.43) {UF};
\fill[secondary!80] (68,-0.55) rectangle (73,-0.31);
\node[font=\scriptsize,anchor=west] at (75,-0.43) {DF};
\end{tikzpicture}
\end{minipage}
\hfill
\begin{minipage}[c]{0.43\linewidth}
\centering
\small
\setlength{\tabcolsep}{4pt}
\begin{tabular}{@{}lrr@{}}
\toprule
Measure & Human & Sol \\
\midrule
Prior rotations & 2.40 & 4.18 \\
Prior coverage (\%) & 42.01 & 46.21 \\
$\Delta$IoU & 0.0207 & 0.0071 \\
$-\Delta$pose ($^\circ$) & 10.0420 & 2.8740 \\
\bottomrule
\end{tabular}
\end{minipage}
\caption{Context and action outcomes across 211 matched states. Prior rotations and coverage are measured before each matched action. UC, DC, UF, and DF denote the four combinations of IoU increasing or decreasing and pose error decreasing or increasing, respectively.}
\label{fig:shared-state-context}
\end{figure*}

\subsection{Action Association with Projection and Pose Discrepancies}
\label{sec:results-guidance}

We next examine whether rotation-angle choices rely more on visible projection overlap or on the underlying 3D pose discrepancy. Figure~\ref{fig:all-method-correlations} and Appendix Table~\ref{tab:all-method-correlations} report their correlations with pre-action IoU and pose error. For GPT-5.6-Sol, the correlation is stronger with IoU than with pose error ($\rho=-0.536$, 95\% CI $[-0.617,-0.443]$ vs.\ $\rho=0.368$, 95\% CI $[0.166,0.532]$). Human angles are strongly associated with both ($\rho=-0.696$, 95\% CI $[-0.763,-0.633]$, and $\rho=0.751$, 95\% CI $[0.675,0.828]$). Most evaluated models show the same relative pattern as GPT-5.6-Sol. These results suggest that most models rely more on projection overlap when choosing rotation angles, whereas humans use both projection overlap and 3D pose discrepancy.

To conduct a comparative analysis under the same conditions as closely as possible, we examine action consequences across the 211 exact state pairs shared by Human and GPT-5.6-Sol, where direct visual coverage before the matched actions is close (42.01\% vs.\ 46.21\%, Figure~\ref{fig:shared-state-context} and Appendix Table~\ref{tab:shared_state_context}). We find that image-space and pose-space progress can diverge even when actions start from the same pose. Humans execute substantially more DC actions (33.18\% vs.\ 23.70\%), showing a willingness to accept transient drops in 2D overlap to reduce underlying 3D pose error. Conversely, GPT-5.6-Sol executes nearly twice as many UF actions as humans (19.43\% vs.\ 10.90\%), showing a greater willingness to accept increases in 3D pose error to improve 2D overlap. For the remaining outcomes, humans achieve far higher UC actions (50.24\% vs.\ 34.60\%) and avoid DF moves (5.69\% vs.\ 22.27\%). Together, these results show that image-space and pose-space progress can diverge: at matched states, GPT-5.6-Sol more often improves silhouette overlap while worsening pose alignment than humans.

\subsection{Limitations and Future Work}
\label{sec:results-limitations}

This work has several limitations. First, the 3D objects are extruded from a limited set of 2D icons, which constrains shape diversity and cannot fully represent complex 3D geometries. Second, each task features only a single target under discrete rotation actions, making the interaction dynamics relatively simplified compared to continuous physical environments. Future work can expand task and icon diversity by leveraging generative models to synthesize richer 2D icons or directly construct more expressive 3D meshes. In addition, the benchmark can be extended by introducing continuous rotation control, external occlusions, and multi-object alignment or assembly tasks to further heighten the spatial reasoning challenge. Finally, automatically generated tasks could provide scalable environments for reinforcement learning fine-tuning, offering a way to study whether interactive training improves the use of visual feedback for precise 3D control.

\section{Conclusion}
\label{sec:conclusion}

RoTA-Bench evaluates whether vision-language models can turn 2D observations into 3D understanding that supports precise action through interactive projection alignment. Its automated pipeline constructs silhouette-preserving 3D objects and uses reversible initialization to guarantee solvability, yielding 674 tasks and a stratified 100-task test set. Evaluations reveal a substantial human--model success gap, show that image-space and pose-space progress can diverge, and find that fine-grained control is sensitive to interaction history. These findings highlight the challenge of converting visual evidence into precise 3D control.

\subsection*{AI use statement}

Generative AI tools were used to aid and polish the writing and to retrieve and discover related work. The vision-language models evaluated in this paper served solely as experimental subjects and were not used to assist the authors' research process. The authors reviewed all AI-assisted materials, verified the reported claims and experimental results, and take full responsibility for the final content of this paper.

\subsection*{Ethics Statement}
All human evaluation experiments were conducted in accordance with ethical research standards, with informed consent obtained from all 20 undergraduate participants prior to the study. Participants were briefed on the task and received fair financial compensation for their time. The evaluation platform displayed only task interactions and recorded only discrete action selections and interaction timestamps, without collecting any personally identifiable information (PII) or sensitive personal data. Furthermore, all 3D task geometries were procedurally constructed from open-source 2D vector icons under valid Creative Commons Attribution 3.0 (CC BY 3.0) and CC0 public domain licenses, with complete author attributions preserved.
\subsection*{Reproducibility Statement}
The release provides the task construction code, evaluation pipeline, and the complete 674-task dataset. The appendix details full prompting protocols, 3D geometric construction, and validation pipelines.
% 参考文献自然紧随其后

% 附录另起新页
\clearpage
\appendix
\makeatletter
\setlength{\@fptop}{0pt}
\setlength{\@dblfptop}{0pt}
\makeatother
\section{Appendix}

The appendix covers icon sources and licensing, human evaluation, target orientation and icon details, icon processing and observation generation, search reference construction, dataset composition and test selection, implementation and evaluation settings, prompt examples, and success and failure examples.

\section{Icon Sources and Licensing}
\label{app:icon-sources}
RoTA-Bench uses SVG shape assets from the official Game Icons repository
(``https://github.com/game-icons/icons''), using the repository snapshot at
commit ``b659ba7773ed6ad1e1fee228b089f7104bea4984'' (May 4, 2025) that was
available during task construction. The repository's ``license.txt'' states
that icons are provided under the Creative Commons Attribution 3.0 license,
or under CC0 when explicitly indicated. We preserve the upstream attribution
and license information with the source assets. The SVG files are rasterized,
quality-filtered, converted to geometric point sets, and rendered as grayscale
three-dimensional objects; the original source colors are not used.

The final canonical collection contains 674 tasks and 674 unique source
icons. Of these, 641 icons are from contributor directories identified as CC
BY 3.0, 33 are from the explicitly marked CC0 directories
(``viscious-speed'' and ``zeromancer'').

Icons made by Lorc, Delapouite, Aussiesim, Viscious Speed, Sbed, Skoll,
DarkZaitzev, Caro Asercion, Cathelineau, Lucas, Andy Meneely, Carl Olsen,
Zeromancer, Guard13007, John Redman, SpencerDub, Felbrigg, Sparker,
Lord Berandas, GeneralAce135, and Faithtoken. The source license is
available at \url{https://github.com/game-icons/icons/blob/b659ba7773ed6ad1e1fee228b089f7104bea4984/license.txt};
license terms are available at \url{https://creativecommons.org/licenses/by/3.0/}
and \url{https://creativecommons.org/publicdomain/zero/1.0/}.

\begin{table}[!t]
\centering\small
\caption{Test-task counts and GPT-5.6-Sol successes by Search-reference length, with success rates for the 4--6-step and 7--16-step groups.}
\label{tab:search-difficulty}
\begin{tabular}{llrrr}
\toprule
Group & Search steps & Tasks & \shortstack{GPT-5.6-Sol\\successes} & \shortstack{GPT-5.6-Sol\\success rate} \\
\midrule
\multirow{3}{*}{4--6 steps} & 4 & 8 & 3 & \multirow{3}{*}{12/55 (21.8\%)} \\
 & 5 & 19 & 3 & \\
 & 6 & 28 & 6 & \\
\midrule
\multirow{10}{*}{7--16 steps} & 7 & 19 & 2 & \multirow{10}{*}{5/45 (11.1\%)} \\
 & 8 & 11 & 1 & \\
 & 9 & 6 & 2 & \\
 & 10 & 3 & 0 & \\
 & 11 & 1 & 0 & \\
 & 12 & 2 & 0 & \\
 & 13 & 1 & 0 & \\
 & 14 & 1 & 0 & \\
 & 15 & 0 & 0 & \\
 & 16 & 1 & 0 & \\
\bottomrule
\end{tabular}
\end{table}

\section{Human Evaluation}
\label{app:human-evaluation}

\paragraph{Participants and formal evaluation.}
Twenty undergraduate participants each completed one consecutive 20-task block in the fixed order of the 100-task test set. The same test set was completed four times, yielding 400 successful human task runs; each participant achieved 100\% success on their assigned tasks. Participation was voluntary, with informed consent and fair compensation. No personally identifiable information was collected. Future work will evaluate participants beyond the undergraduate population.

\paragraph{Interface and procedure.}
Participants received no practice trials or additional instruction before the formal evaluation. At every step, they received the same target projection, auxiliary view, current projection, and chronological history information as the evaluated models. The task environment, observation content, action vocabulary, allowed angles, success threshold, stopping rules, and pose-transition function were identical for humans and models. Humans selected the same discrete actions and angles through interface buttons, while models returned the corresponding actions in JSON. Thus, the interaction space and environment input--output mapping were identical; the buttons only provided a convenient human control interface. Human reasoning was not recorded.

\paragraph{Response validity and time.}
The model evaluator obtained valid JSON responses for all logged actions within at most six retries. This confirms that the structured action interface was available throughout evaluation; human button selections represented the same action tokens and angles. Across the 400 human task runs, 99\% were completed within 278.12 seconds (4.64 minutes) per task. Task duration includes the time between task initialization and the final action, including pauses between actions.

\paragraph{Individual results and aggregation.}
Table~\ref{tab:human-individual} reports individual results. Task metrics average the four human task runs for each test task, then average across tasks. Action metrics combine all executed rotations; angle metrics use only effective directions.

\begin{table}[htbp]
\centering
\small
\setlength{\tabcolsep}{4pt}
\caption{Shared human and model evaluation protocol. The interaction environment, visible observations, action space, and task rules are identical; humans use buttons to submit the same actions that models return in JSON.}
\label{tab:human-model-protocol}
\begin{tabular}{@{}p{0.27\linewidth}p{0.63\linewidth}@{}}
\toprule
Setting & Shared protocol \\
\midrule
Tasks & Same 100-task test set; each participant completes one consecutive 20-task block \\
Current observations & Target projection, auxiliary view, current projection, and chronological history \\
Rotation actions & Six signed directions with $1^\circ$, $5^\circ$, $10^\circ$, or $30^\circ$ \\
Action representation & Human button selection and model JSON both map to the same action token and angle \\
Success criterion & Projection IoU $\geq 0.90$ \\
Stopping rules & Success, STOP, 48 decisions, or 12 rotations without a new best IoU \\
Hidden information & Numerical IoU, internal poses, and reference paths remain hidden \\
Environment transition & The same pose update and observation regeneration after every action \\
\bottomrule
\end{tabular}
\end{table}

\begin{table}[htbp]
\centering
\small
\caption{Human results for four passes over the 100-task test set. Each participant completed one consecutive 20-task block; participants with the same task-block index evaluated the same 20 tasks in different passes.}
\label{tab:human-individual}
\begin{tabular}{lrrrrrr}
\toprule
Participant & Pass & Task block & Successes/tasks & Success (\%) & Final IoU & Rotations \\
\midrule
1  & 1 & 1 & 20/20 & 100.00 & 0.9198 & 13.05 \\
2  & 1 & 2 & 20/20 & 100.00 & 0.9190 & 11.90 \\
3  & 1 & 3 & 20/20 & 100.00 & 0.9162 & 13.15 \\
4  & 1 & 4 & 20/20 & 100.00 & 0.9172 & 12.05 \\
5  & 1 & 5 & 20/20 & 100.00 & 0.9246 & 13.20 \\
6  & 2 & 1 & 20/20 & 100.00 & 0.9283 & 18.85 \\
7  & 2 & 2 & 20/20 & 100.00 & 0.9285 & 17.40 \\
8  & 2 & 3 & 20/20 & 100.00 & 0.9266 & 14.95 \\
9  & 2 & 4 & 20/20 & 100.00 & 0.9290 & 18.95 \\
10 & 2 & 5 & 20/20 & 100.00 & 0.9265 & 16.65 \\
11 & 3 & 1 & 20/20 & 100.00 & 0.9149 & 16.50 \\
12 & 3 & 2 & 20/20 & 100.00 & 0.9197 & 14.25 \\
13 & 3 & 3 & 20/20 & 100.00 & 0.9223 & 14.10 \\
14 & 3 & 4 & 20/20 & 100.00 & 0.9233 & 16.55 \\
15 & 3 & 5 & 20/20 & 100.00 & 0.9205 & 16.05 \\
16 & 4 & 1 & 20/20 & 100.00 & 0.9179 & 23.15 \\
17 & 4 & 2 & 20/20 & 100.00 & 0.9259 & 17.75 \\
18 & 4 & 3 & 20/20 & 100.00 & 0.9283 & 16.80 \\
19 & 4 & 4 & 20/20 & 100.00 & 0.9229 & 20.70 \\
20 & 4 & 5 & 20/20 & 100.00 & 0.9252 & 18.35 \\
\midrule
Combined & -- & -- & 400/400 & 100.00 & 0.9228 & 16.22 \\
\bottomrule
\end{tabular}
\end{table}

\begin{figure}[!t]
\centering
\includegraphics[width=\linewidth]{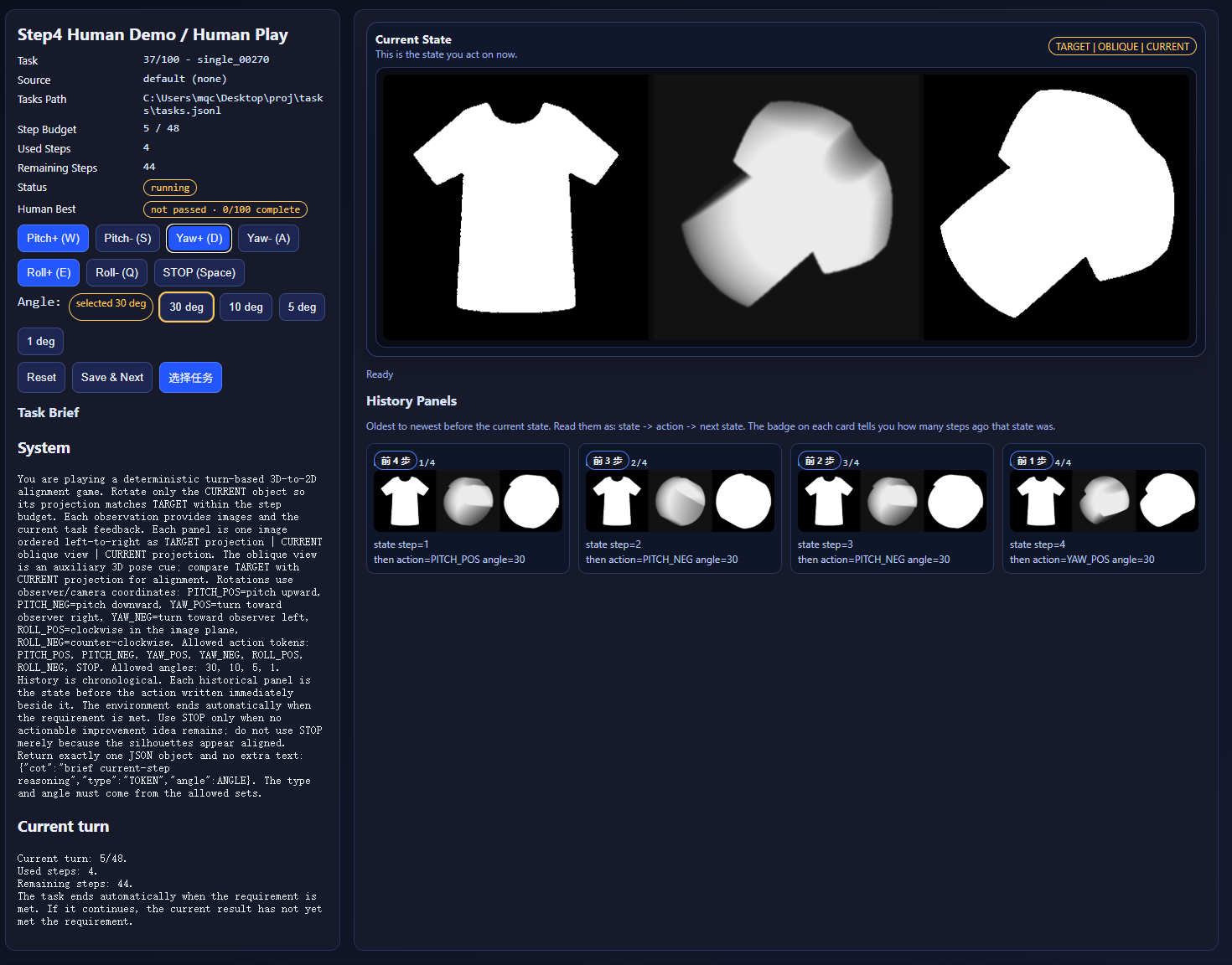}
\caption{Human evaluation interface with the same prompt content, current observation, historical observations, and discrete action space used by the models. Humans submit actions through buttons that map to the model action tokens and angles.}
\label{fig:human-web-demo}
\end{figure}
\FloatBarrier

\section{Computation of Object Symmetries}
\label{app:object-symmetries}
Symmetry allows different orientations to represent the same object pose. We estimate these symmetries from the normalized 2D points used to build each object, without using evaluation results.

\paragraph{Finding symmetries.}
We test rotational orders $n=12,11,\ldots,2$. Order $n$ requires every rotation by $360^\circ k/n$, $k=1,\ldots,n-1$, to pass the point-matching check below. We keep the first accepted order; otherwise, we keep only the identity rotation. For reflections, we test $Q(\alpha)=\operatorname{Rot}(\alpha)\operatorname{diag}(1,-1)$, where $\alpha$ is twice the mirror-axis angle. We search $0^\circ\leq\alpha<360^\circ$ in $2^\circ$ steps and refine the four best candidates within $\pm2^\circ$. A reflection and all its combinations with the retained rotations must pass the same check.

\paragraph{Checking point matches.}
A symmetry should map the point set onto itself. We allow a small error because the points are sampled. Using up to 1,500 evenly indexed points, we estimate the median distance $h$ to the nearest other point in the full set and set $\tau=3h+10^{-6}$. We measure nearest-neighbor distances in both directions between the original and transformed sets. In each direction, the 99th-percentile distance must be at most $\tau$, and the maximum must be at most $2\tau$. We screen candidates with the sample, then check accepted candidates with all points.

\paragraph{Converting to 3D.}
The object's depth is symmetric about the 2D plane and depends only on distance from the center. Each planar symmetry $Q$ therefore gives a 3D rotation
\begin{equation}
S(Q)=\begin{pmatrix}Q&0\\0&\det Q\end{pmatrix}.
\end{equation}
Planar rotations preserve depth; planar reflections also reverse depth. Both give determinant $+1$. These matrices, including the identity, form $\mathcal{G}$. We check that each matrix product matches a retained matrix within $10^{-6}$ in Frobenius norm.

\paragraph{Measuring pose error.}
Equation~\ref{eq:pose-error} takes the smallest rotation distance from the current pose to any equivalent target $R^\star S$. We clip the arccosine input to $[-1,1]$ and report degrees. This is a numerical symmetry estimate limited by sampling tolerance and the search range. A matching silhouette need not imply an equivalent 3D pose, so pose error is a diagnostic and IoU determines success.

\section{Target Orientation and Icon Details}
\label{app:target-orientation}
We set $R_{\mathrm{target}}=I_3$ because projection along the depth axis retains the source icon's details. For the continuous construction in Equation~\ref{eq:depth-envelope}, projecting $(x,y,z)$ to $(x,y)$ recovers $P$: depth extension changes $z$ but preserves the planar coordinates. Sampling and rasterization approximate this relation. At other orientations, extended depth can overlap in projection and obscure these details; the target need not have a unique matching orientation. Figure~\ref{fig:case-success} illustrates this distinction: the initial projection is nearly circular, whereas the target preserves the shirt contour.

\section{Surface Coverage}
\label{app:surface-coverage}

We measure how much of an object becomes visible during an episode. First, we reconstruct the object's closed triangular surface with a grid spacing of $0.75$ and assign each triangle its geometric area. We use the triangle centroid as its representative surface point. At each observed pose, a point is marked visible if its outward normal faces the camera and a ray from the point to the camera is not blocked by another triangle. Visibility is tested with exact ray--triangle intersections using a bounding-volume hierarchy.

Let $V_t$ denote the visible surface points at step $t$, and let $w_i$ be the area weight of point $i$. We accumulate visibility over the complete trajectory:
\begin{equation}
V_{\mathrm{episode}}=\bigcup_{t=0}^{T} V_t,
\qquad
\mathrm{Coverage}=
\frac{\sum_{i\in V_{\mathrm{episode}}} w_i}
{\sum_i w_i}.
\end{equation}
Thus, a surface point is counted once even if it is observed at multiple steps. The reported coverage is the area-weighted fraction of the reconstructed surface observed at least once during the episode.

\section{Icon Processing and Observation Generation}
\label{app:icon-processing}

Rasterization and foreground extraction. We rasterize each SVG to $512\times512$ RGBA pixels and estimate the background color $c_{\rm bg}$ from the four corners. For pixel $(u,v)$, $C(u,v)$ is its RGB vector and $\alpha(u,v)$ its opacity. We retain a pixel when it is sufficiently opaque and differs from the background:
\begin{equation}
F(u,v)=\mathbf{1}[\alpha(u,v)>8\ \land\ \|C(u,v)-c_{\rm bg}\|_1>30].
\label{eq:spar-foreground}
\end{equation}
Here, $\mathbf{1}$ is an indicator, $\land$ requires both conditions, and $\|\cdot\|_1$ sums absolute RGB differences. This separates icon content from transparency and background before geometry construction.

Point extraction and filtering. We scan foreground pixels row by row and map $(u,v)$ to $(24(u-256)/256,24(v-256)/256)$, rounded to four decimals. If there are $n>90{,}000$ points, we keep indices $\lfloor k(n-1)/(m-1)\rfloor$ for $m=90{,}000$ and $k=0,\ldots,m-1$; otherwise, all points are kept. This deterministic sampling bounds computation while preserving the icon outline. On a $112\times112$ occupancy grid, we retain icons only if the largest-component ratio is at least 0.985, circle occupancy at least 0.28, bounding-box occupancy at least 0.08, aspect ratio at most 2.2, and foreground count at least 2,000. These tests remove fragmented or degenerate shapes that would produce ambiguous tasks; circle occupancy is an area estimate, not mask IoU.

Geometry construction. We multiply points by 1.24, center them at their minimum enclosing circle, and fit them within $0.98r$ for $r=28$. For centered $(x,y)$, the sphere-bounded half-depth is $h(x,y)=\sqrt{\max(r^2-x^2-y^2,0)}$ (Equation~\ref{eq:depth-envelope}); the maximum prevents an invalid square root. Depth and complexity scales are 1. We use 31 samples per interval: endpoints $z_{i0}=-h(x_i,y_i)$ and $z_{i1}=h(x_i,y_i)$, plus 29 deterministic pseudorandom interior samples:
\begin{equation}
z_{ij}=(2q_{ij}-1)h(x_i,y_i),\qquad j=2,\ldots,30.
\end{equation}
Here, $i$ and $j$ are point and sample indices, $q_{ij}\in\left[0,1\right)$ is a deterministic function of these indices and the task seed, and $2q_{ij}-1$ maps it into the depth interval. This gives the 2D outline finite 3D thickness while keeping regeneration deterministic; the seed is stored with the task.

Projection. A rotated point $(x_t,y_t,z_t)$ is mapped orthographically to $(c_x+\lambda x_t,c_y+\lambda y_t)$; $z_t$ is depth, $(c_x,c_y)=(160,160)$, and $\lambda=5$ for $320\times320$ images. Coordinates are rounded, a 2-pixel disk is marked (255), and out-of-bounds pixels are discarded. Orthographic projection keeps alignment dependent on the visible outline rather than depth-dependent scale.

Auxiliary image. To add depth cues, the same point is mapped to $(c_x+0.9\lambda x_t,c_y+0.9\lambda y_t)$ with intensity
\begin{equation}
g_t=\min(235,\max(40,110+6z_t+1.8x_t)).
\end{equation}
Here, 0.9 is the scale factor, 110 the baseline, 6 and 1.8 control depth and horizontal-position effects, and $\min(235,\max(40,\cdot))$ clips intensity to $[40,235]$. The resulting gradient supplies depth cues that the binary mask does not contain.

Image smoothing. For Game-icons at $320\times320$, each point is rasterized with radius $\operatorname{round}[(320/64)\times0.36]=2$ pixels (the code computes this from image scale and source type), filling small gaps without strongly expanding the outline. The auxiliary background is 18; within each disk, intensity decreases by $\lfloor2.2\sqrt{\Delta u^2+\Delta v^2}\rfloor$, and overlaps keep the maximum. Both masks receive one morphological closing with a $3\times3$ elliptical kernel, and the auxiliary image receives two $3\times3$ Gaussian passes ($\sigma=0$). These steps reduce raster gaps; both views use the same orientation.

Observation assembly. The target projection, current auxiliary image, and current projection form a $972\times320$ panel, separated by two 6-pixel strips of intensity 12. This fixed layout gives every method the same visual information. Formal evaluation uses the CUDA renderer with float32 geometry; software versions are listed in Appendix~\ref{app:implementation}. The complete implementation, including deterministic depth sampling, task seeds, and rendering parameters, is provided in the open-source code.
\section{Search Reference Construction}
\label{app:search-reference}
The inverse initialization sequence provides a feasible solution independently of search. Replaying it checks that the implemented transformations and projection scoring reproduce successful alignment. We combine pose-guided Beam-256 search with a breadth-first search (BFS) restricted to $30^\circ$ rotations, following the standard discrete-state search formulation \citep{hart1968formal}. At depth $d$, Beam-256 computes the pose residual for each of the six directions, selects the closest angle from the four allowed angles for each direction, and therefore generates six candidate actions per retained path. It then keeps at most 256 candidates ranked by pose distance to the target and orientation diversity:
\begin{equation}
\mathcal{B}_{d+1}=\mathrm{Top}_{256}\!\left(\lbrace p\,a\mid p\in\mathcal{B}_d,\ a\in\mathcal{A}\rbrace\right),\qquad |\mathcal{A}|=6.
\label{eq:spar-beam}
\end{equation}
The fixed-angle search instead performs BFS over rotation-matrix states, expanding every state at depth $d$ with the same six actions and stopping at the first depth whose replayed projection reaches the success threshold:
\begin{equation}
\mathcal{S}_{d+1}=\lbrace R(a)R\mid a\in\mathcal{A},\ R\in\mathcal{S}_d\rbrace,\qquad \mathcal{S}_0=\lbrace R_{\rm init}\rbrace.
\label{eq:spar-bfs}
\end{equation}
Both searches replay candidate paths and verify them with the same projection IoU threshold used in evaluation. We retain the shorter verified candidate as the Search reference, although it is not necessarily optimal. Restricting the search keeps reference construction computationally feasible, while internal poses and reference paths remain hidden from evaluated models, agents, and humans. The complete implementation is provided in the open-source code.

Table~\ref{tab:reference-length-comparison} summarizes the comparison between the inverse initialization sequence and the resulting Search reference over all 674 canonical tasks.

\begin{table}[!t]
\centering
\small
\caption{Search-reference and inverse-initialization path lengths across 674 tasks. Counts and percentages compare the paired lengths; summary statistics report the mean, median, and range in rotation steps.}
\label{tab:reference-length-comparison}
\begin{tabular}{lrr}
\toprule
Comparison & Tasks & Percentage \\
\midrule
Search reference shorter & 585 & 86.8\% \\
Equal length & 89 & 13.2\% \\
Search reference longer & 0 & 0.0\% \\
\midrule
\multicolumn{3}{l}{Inverse mean / median / range: 11.04 / 11 / 6--16} \\
\multicolumn{3}{l}{Search mean / median / range: 6.54 / 6 / 2--16} \\
\bottomrule
\end{tabular}
\end{table}

\section{Dataset Composition and Test Selection}
\label{app:dataset-construction}
The canonical collection contains 674 tasks with distinct source-icon identifiers. The 642 tasks whose reference paths contain at least four rotations are eligible for test selection. We allocate 100 test places proportionally across reference-length groups using the Hamilton largest-remainder method: first assign the integer part of each proportional quota, then assign remaining places in descending order of fractional remainder. Within each group, tasks are ranked deterministically by a SHA-256 hash of the selection seed and task identifier. Table~\ref{tab:spar-dataset-strata} gives the full dataset and test-set counts. The remaining 574 tasks are reserved for future training and research.

\begin{table}[!t]
\centering
\small
\caption{Task counts in the 674-task dataset and 100-task test set by Search-reference length. Tasks with at least four reference rotation steps are eligible for test selection.}
\label{tab:spar-dataset-strata}
\begin{tabular}{lrr}
\toprule
Reference rotations & All tasks & Test tasks \\
\midrule
$<4$ & 32 & 0 \\
4 & 53 & 8 \\
5 & 122 & 19 \\
6 & 181 & 28 \\
7 & 124 & 19 \\
8 & 69 & 11 \\
9 & 38 & 6 \\
10 & 21 & 3 \\
11 & 9 & 1 \\
12 & 10 & 2 \\
13 & 5 & 1 \\
14 & 5 & 1 \\
15 & 1 & 0 \\
16 & 4 & 1 \\
\midrule
Eligible ($\geq4$) & 642 & 100 \\
Total & 674 & 100 \\
\bottomrule
\end{tabular}
\end{table}

\section{Additional Results}
\label{app:additional-results}

\begin{table*}[t]
\centering
\small
\setlength{\tabcolsep}{4pt}
\caption{Context before matched actions and their outcomes across 211 participant--task--state pairs. Prior rotations count completed rotations before each action. Prior coverage is cumulative direct surface coverage up to and including the current observation. All values are averaged over pairs. UC improves IoU and reduces pose error; DC reduces pose error only; UF improves IoU only; DF improves neither.}
\label{tab:shared_state_context}
\begin{tabular}{@{}lcccccccc@{}}
\toprule
Method & \shortstack{Prior\\rotations} & \shortstack{Prior\\coverage}
& $\Delta$IoU & UC & DC & UF & DF & $-\Delta$pose ($^\circ$) \\
\midrule
Human & 2.40 & 42.01\% & 0.0207 & 50.24\% & 33.18\% & 10.90\% & 5.69\% & 10.0420 \\
GPT-5.6-Sol & 4.18 & 46.21\% & 0.0071 & 34.60\% & 23.70\% & 19.43\% & 22.27\% & 2.8740 \\
\bottomrule
\end{tabular}
\end{table*}

% Appendix supplementary table
\begin{table*}[t]
\centering
\small
\setlength{\tabcolsep}{4pt}
\caption{Additional pose and trajectory statistics on the same 100 tasks. Final pose error is the symmetry-adjusted angular distance from the target at the end of a trajectory. Pose-improving steps are the percentage of all executed steps that reduce pose error. Mean $\Delta$IoU per step and mean pose reduction per step are computed over complete trajectories; positive values indicate progress toward the target. Human combines 400 task runs from 20 undergraduate participants, and random policies average three seeds within each task. SpatioLM-InternVL3.5 abbreviates SpatioLM-Understanding-InternVL3.5.}
\label{tab:supplementary-pose-results}
\begin{tabular}{@{}lrrrr@{}}
\toprule
Method
& \shortstack{Final pose\\error ($^\circ$)}
& \shortstack{Pose-impr.\\steps (\%)}
& \shortstack{Mean $\Delta$IoU\\/ step}
& \shortstack{Mean pose red.\\/ step ($^\circ$)} \\
\midrule
Human                     &   4.44 &  77.88 &  0.0273 &  5.4238 \\
IoU oracle                &  79.06 &  54.82 &  0.0280 &  2.8687 \\
Pose oracle               &   1.67 & 100.00 &  0.0535 & 10.9053 \\
Uniform random            & 102.11 &  48.94 & -0.0003 & -0.4365 \\
Random + IoU rollback     &  88.83 &  52.14 &  0.0018 &  0.1002 \\
ReAct                     &  82.69 &  52.84 &  0.0060 &  0.3944 \\
Gemini 3.6 Flash          &  60.88 &  53.57 &  0.0052 &  1.2771 \\
GPT-5.6-Sol               &  85.25 &  52.16 &  0.0050 &  0.3021 \\
GPT-6-Astra               &  84.50 &  50.09 &  0.0043 &  0.3408 \\
SpatioLM-InternVL3.5      & 111.35 &  40.72 & -0.0009 & -1.0074 \\
GPT-5.6-Terra             &  92.48 &  50.28 &  0.0033 & -0.0035 \\
GPT-5.6-Luna              &  93.66 &  50.80 &  0.0030 & -0.0518 \\
Magentic                  &  91.49 &  50.50 &  0.0031 &  0.0566 \\
Qwen 3.7 Plus             &  93.32 &  47.84 &  0.0012 & -0.0440 \\
Claude Sonnet 5           & 106.91 &  46.88 &  0.0006 & -0.7482 \\
Claude Opus 5             & 102.13 &  47.62 &  0.0002 & -0.4403 \\
\bottomrule
\end{tabular}
\end{table*}

\FloatBarrier
\section{Additional Evidence for the Three Findings}
\label{app:finding-evidence}

% =========================================================================
% 3. 相关性表 (单栏排版，方法排序严格与 Table 1 的 Final IoU 一致)
% =========================================================================
\begin{table}[t]
\centering
\small
\setlength{\tabcolsep}{2.5pt}
\caption{Spearman rank correlations ($\rho$) of selected rotation angles with pre-action projection IoU (IoU--angle) and 3D pose error (Pose--angle) across the 100 test tasks, sorted by Final IoU. Brackets report 95\% bootstrap confidence intervals.}
\label{tab:all-method-correlations}
\begin{tabular*}{\linewidth}{@{\extracolsep{\fill}}lcc@{}}
\toprule
Method & IoU--angle $\rho$ (95\% CI) & Pose--angle $\rho$ (95\% CI) \\
\midrule
Human             & $\mathbf{-0.696}$ [$-0.763, -0.633$] & $\mathbf{0.751}$ [$0.675, 0.828$] \\
\midrule
ReAct             & $-0.570$ [$-0.674, -0.453$]          & $0.320$ [$0.108, 0.512$] \\
Gemini 3.6 Flash  & $-0.562$ [$-0.633, -0.480$]          & $0.618$ [$0.496, 0.709$] \\
GPT-5.6-Sol       & $-0.536$ [$-0.617, -0.443$]          & $0.368$ [$0.166, 0.532$] \\
GPT-6-Astra       & $-0.533$ [$-0.621, -0.427$]          & $0.452$ [$0.266, 0.607$] \\
GPT-5.6-Terra     & $-0.438$ [$-0.526, -0.335$]          & $0.254$ [$0.056, 0.440$] \\
GPT-5.6-Luna      & $-0.471$ [$-0.561, -0.373$]          & $0.280$ [$0.088, 0.470$] \\
Magentic          & $-0.368$ [$-0.493, -0.215$]          & $0.250$ [$0.066, 0.403$] \\
Qwen 3.7 Plus     & $-0.383$ [$-0.495, -0.251$]          & $0.380$ [$0.232, 0.514$] \\
Claude Sonnet 5   & $-0.255$ [$-0.365, -0.140$]          & $0.154$ [$-0.021, 0.309$] \\
Claude Opus 5     & $-0.091$ [$-0.229,  0.044$]          & $0.204$ [$0.043, 0.354$] \\
SpatioLM          &  $0.127$ [$-0.063,  0.309$]          & $-0.142$ [$-0.268, 0.005$] \\
\bottomrule
\end{tabular*}
\end{table}

\subsection{Success Counts and Confidence Intervals}
\label{app:success-intervals}
Table~\ref{tab:success-intervals} reports success counts and 95\% Wilson confidence intervals for the nine models and two agents on the 100-task test set. These intervals summarize uncertainty under a binomial task model. GPT-5.6-Sol has the highest observed success rate, but this ranking alone does not establish a reliable difference between methods. Such comparisons require paired analysis on the same tasks.

\begin{table}[htbp]
\centering
\small
\setlength{\tabcolsep}{6pt}
\caption{Success counts and 95\% Wilson confidence intervals for models and agents on the 100 test tasks. Rows follow the order in Table~\ref{tab:formal-results}.}
\label{tab:success-intervals}
\begin{tabular}{@{}lrrc@{}}
\toprule
Method & Successes/tasks & Success (\%) & 95\% CI (\%) \\
\midrule
ReAct & 16/100 & 16.00 & [10.10, 24.42] \\
Gemini 3.6 Flash & 10/100 & 10.00 & [5.52, 17.44] \\
GPT-5.6-Sol & 17/100 & 17.00 & [10.89, 25.55] \\
GPT-6-Astra & 10/100 & 10.00 & [5.52, 17.44] \\
GPT-5.6-Terra & 8/100 & 8.00 & [4.11, 15.00] \\
GPT-5.6-Luna & 4/100 & 4.00 & [1.57, 9.84] \\
Magentic & 2/100 & 2.00 & [0.55, 7.00] \\
Qwen 3.7 Plus & 0/100 & 0.00 & [0.00, 3.70] \\
Claude Sonnet 5 & 0/100 & 0.00 & [0.00, 3.70] \\
Claude Opus 5 & 0/100 & 0.00 & [0.00, 3.70] \\
SpatioLM-InternVL3.5 & 0/100 & 0.00 & [0.00, 3.70] \\
\bottomrule
\end{tabular}
\end{table}

\subsection{Response Formats and Episode Outcomes}
\label{app:failure-log-audit}

We audit the saved trajectories of all nine models and two agents on the 100 test tasks each. Table~\ref{tab:failure-outcomes} reports episode outcomes and repeated actions. Stagnation denotes termination after consecutive rotations without sufficient improvement in the best IoU. STOP denotes an explicit stop below the success threshold, and Limit denotes termination at the step limit. Repeat is the percentage of consecutive rotation pairs with the same signed axis and angle within an episode. Repeated actions can be useful, and stopping reasons alone do not explain why a task failed.

\begin{table*}[t]
\centering
\small
\setlength{\tabcolsep}{5pt}
\caption{Episode outcomes and repeated actions on 100 tasks per method. The four outcome columns sum to 100. Repeat is measured over consecutive rotation pairs.}
\label{tab:failure-outcomes}
\begin{tabular*}{\linewidth}{@{\extracolsep{\fill}}lrrrrr@{}}
\toprule
Method & Success & Stagnation & STOP & Limit & Repeat (\%) \\
\midrule
Human-independent model methods & & & & & \\
GPT-5.6-Sol & 17 & 78 & 3 & 2 & 53.35 \\
ReAct & 16 & 75 & 5 & 4 & 49.41 \\
Gemini 3.6 Flash & 10 & 83 & 2 & 5 & 23.90 \\
GPT-6-Astra & 10 & 85 & 1 & 4 & 53.65 \\
GPT-5.6-Terra & 8 & 84 & 2 & 6 & 53.90 \\
GPT-5.6-Luna & 4 & 91 & 3 & 2 & 52.13 \\
Magentic & 2 & 44 & 52 & 2 & 23.32 \\
Qwen 3.7 Plus & 0 & 99 & 1 & 0 & 38.34 \\
Claude Sonnet 5 & 0 & 95 & 5 & 0 & 23.60 \\
Claude Opus 5 & 0 & 95 & 3 & 2 & 19.16 \\
SpatioLM & 0 & 100 & 0 & 0 & 100.00 \\
\bottomrule
\end{tabular*}
\end{table*}

\begin{table*}[t]
\centering
\small
\setlength{\tabcolsep}{5pt}
\caption{Distribution of successfully returned actions by the number of retries needed to obtain valid JSON, over saved trajectories. Only responses that contain output tokens count as valid returns; transmission failures (no returned tokens) are excluded. ``retries$\leq k$'' is the cumulative share of actions produced within $k$ retries (i.e.\ on the $k{+}1$-th content-bearing response).}
\label{tab:response-formats}
\begin{tabular*}{\linewidth}{@{\extracolsep{\fill}}lrrrr@{}}
\toprule
Method & Legal JSON & retries$=$0 (\%) & $\leq$2 (\%) & $\leq$6 (\%) \\
\midrule
GPT-5.6-Sol & 2369 & 99.87 & 100.00 & 100.00 \\
GPT-5.6-Luna & 2432 & 92.27 & 100.00 & 100.00 \\
GPT-5.6-Terra & 2293 & 100.00 & 100.00 & 100.00 \\
Gemini 3.6 Flash & 2470 & 99.92 & 100.00 & 100.00 \\
Claude Opus 5 & 2212 & 99.55 & 100.00 & 100.00 \\
Claude Sonnet 5 & 1944 & 100.00 & 100.00 & 100.00 \\
Qwen 3.7 Plus & 2104 & 99.95 & 100.00 & 100.00 \\
ReAct & 2467 & 99.80 & 100.00 & 100.00 \\
Magentic & 1652 & 99.52 & 100.00 & 100.00 \\
SpatioLM & 1881 & 100.00 & 100.00 & 100.00 \\
GPT-6-Astra     & 2319 & 90.86 & 98.45 & 100.00  \\
\bottomrule
\end{tabular*}
\end{table*}

We verified, on the saved trajectories, that every logged action was eventually returned as valid JSON, counting only responses that contain output tokens and excluding transmission failures. Ten of the eleven methods returned all actions in valid JSON within at most two retries. GPT-6-Astra had 2,318 rotation actions and one additional content-bearing STOP response; across these 2,319 returns, 90.86\% were valid on the first response, 98.45\% within two retries, and 100\% within six retries.

\subsection{Statistical Units and Resampling}
We use existing trajectories only. Action rates combine all executed rotations; angle metrics use only effective directions. To estimate uncertainty, we repeatedly sample tasks with replacement and keep all steps from each sampled task. For each sampled task, we keep its four human task runs together. Paired comparisons use the same sampled tasks for both methods. For the additional bootstrap analyses in this section, we use 10,000 resamples and report the 2.5th and 97.5th percentiles as the 95\% confidence interval. 

Let $\Delta s$ and $\Delta e$ denote changes in IoU and pose error. UC increases IoU and reduces pose error; DC decreases both; UF increases both; DF decreases IoU and increases pose error. Changes within $10^{-6}$ for IoU or $10^{-4}$ degrees for pose error are treated as neutral.

\subsection{Direction and Angle Diagnostics}
\label{app:finding-one}
Human exceeds Sol in Effective Direction Rate by 25.16 percentage points and has a $3.50^\circ$ smaller Angle Gap (Table~\ref{tab:paired-direction-angle}). Both paired 95\% intervals exclude zero. This supports a difference in observed actions on the same 100 tasks, but does not directly measure the models' internal pose estimates.

\begin{table}[htbp]
\centering\small
\caption{Action metrics on 100 tasks, with paired 95\% bootstrap intervals. Angle Gap uses effective directions only. Differences are Human minus Sol; direction differences are in percentage points.}
\label{tab:paired-direction-angle}
\begin{tabular}{lcc}
\toprule
Method/comparison & Effective Direction Rate (\%) & Angle Gap ($^\circ$) \\
\midrule
Human & 83.91 [82.63, 85.22] & 2.99 [2.70, 3.30] \\
GPT-5.6-Sol & 58.75 [56.62, 60.87] & 6.49 [5.38, 7.64] \\
Human minus Sol & 25.16 [22.79, 27.56] & $-3.50$ [$-4.60$, $-2.42$] \\
\bottomrule
\end{tabular}
\end{table}

The data contain 6,487 Human rotations and 2,366 Sol rotations. Longer episodes contribute more actions, so we resample whole tasks to keep related steps together. 

\subsection{Input Ablations and Fine-Grained Control}
\label{app:finding-two}
We compare four inputs on the same 30 tasks: full input (A), no auxiliary view (B), no history (C), and neither (D). History includes earlier observations, responses, and actions. Removing history keeps the current observation and instructions, but removes all prior context; it does not isolate a single reasoning mechanism.

\begin{table}[htbp]
\centering\small
\caption{Successes on 30 tasks, with Wilson 95\% intervals. The last four columns count rotations at each angle.}
\label{tab:ablation-evidence}
\begin{tabular}{lccrrrr}
\toprule
Input & Successes & Success (\%; 95\% CI) & $1^\circ$ & $5^\circ$ & $10^\circ$ & $30^\circ$ \\
\midrule
A: Full & 5/30 & 16.67 [7.34, 33.56] & 86 & 53 & 124 & 443 \\
B: No auxiliary & 1/30 & 3.33 [0.59, 16.67] & 24 & 22 & 48 & 584 \\
C: No history & 0/30 & 0.00 [0.00, 11.35] & 1 & 3 & 16 & 598 \\
D: Neither & 2/30 & 6.67 [1.85, 21.32] & 8 & 8 & 5 & 597 \\
\bottomrule
\end{tabular}
\end{table}

Without history, $30^\circ$ rotations rise from 443/706 (62.75\%) to 598/618 (96.76\%), while $1^\circ$ rotations fall from 86/706 (12.18\%) to 1/618 (0.16\%). Thus, fine adjustments nearly disappear. Full input also improves success and Final IoU over no history, with paired intervals above zero (Table~\ref{tab:ablation-paired}). The Final IoU interval for A minus B includes zero. Removing both inputs gives two successes, versus zero without history alone; these 30 runs do not explain this reversal.

\begin{table}[htbp]
\centering\small
\caption{Full input minus each ablation on the same 30 tasks, with paired 95\% bootstrap intervals. Success differences are in percentage points.}
\label{tab:ablation-paired}
\begin{tabular}{lcc}
\toprule
Comparison & Success difference & Mean Final IoU difference \\
\midrule
A minus B & 13.33 [0.00, 30.00] & 0.0166 [$-0.0598$, 0.0961] \\
A minus C & 16.67 [3.33, 30.00] & 0.0834 [0.0255, 0.1477] \\
A minus D & 10.00 [0.00, 20.00] & 0.0568 [0.0138, 0.1107] \\
\bottomrule
\end{tabular}
\end{table}

\subsection{Comparing Associations and Shared-State Actions}
\label{app:finding-three}
To compare the two correlations in
Figure~\ref{fig:all-method-correlations}, we compute the difference
in their absolute values:
\begin{equation}
\Delta_\rho =
|\rho(\theta,s_{\mathrm{pre}})|
-
|\rho(\theta,e_{\mathrm{pre}})|.
\end{equation}
Positive values indicate a stronger association with IoU;
negative values indicate a stronger association with pose error.
We recompute both correlations on the same resampled tasks
and obtain a confidence interval for their difference.

For GPT-5.6-Sol, $\Delta_\rho=0.169$
(95\% CI: $[0.020,0.330]$).
For Human, $\Delta_\rho=-0.055$
(95\% CI: $[-0.084,-0.023]$).
These differences use unrounded correlations.
Both intervals exclude zero, supporting a stronger association
with IoU for Sol and with pose error for Human.

\paragraph{Shared-state matching and weighting.}
Table~\ref{tab:shared_state_context} contains 211 participant--task--state pairs from 51 tasks. Within a task, two poses match if their rotation distance is below $10^{-4}$ degrees. We exclude initial states and repeated Human visits to a matched state, then pair the Human action with Sol's first visit. The four test-set passes contribute 52, 20, 64, and 75 pairs. A Sol action may appear in several participant pairs. Each pair has equal weight, and task resampling keeps related pairs together.

Matching controls the current pose, but earlier observations and actions may differ.
Figure~\ref{fig:shared-state-context} reports completed rotations and cumulative
direct coverage before each matched action, averaged over the 211 pairs.
Human reaches these states after 2.40 rotations on average, compared with
4.18 for Sol, with coverage of 42.01\% and 46.21\%, respectively.
These comparisons describe behavior at shared states; they do not establish
a specific internal mechanism or a general limitation in 3D reasoning.

\subsection{Oracle Action Decomposition}
\label{app:oracle-decomposition}

To separate direction and angle selection, we evaluate the 2,366 states visited by GPT-5.6-Sol on the 100-task test set. For each state, we enumerate all 24 legal actions and compute the successor pose error of each action. Here, the best action is defined as the legal action that minimizes the symmetry-adjusted successor pose error. The direction oracle keeps Sol's selected angle and chooses the best rotation axis and sign; the angle oracle keeps Sol's selected direction and chooses the best legal angle; the joint oracle chooses the best action without either constraint. Uniform random averages successor pose error over all 24 actions. As shown in Table~\ref{tab:oracle-decomposition}, correcting direction yields a larger one-step improvement than correcting angle, while correcting both yields the lowest expected error. These results indicate that GPT-5.6-Sol still has substantial room for improvement in both rotation-direction and rotation-angle selection.

\begin{table}[t]
\centering
\small
\setlength{\tabcolsep}{3.5pt}
\caption{Oracle decomposition on 2,366 GPT-5.6-Sol states. Successor pose errors and improvements are averaged over states. For Uniform random, the successor error is the average over all 24 legal actions in each state. Oracle actions are selected by minimizing the symmetry-adjusted successor pose error.}
\label{tab:oracle-decomposition}
\begin{tabular}{@{}lrr@{}}
\toprule
Policy & \shortstack{Successor error\\($^\circ$)}
& \shortstack{Improvement\\($^\circ$)}
\\
\midrule
Sol actual action & 91.99 & 0.00 \\
Uniform random & 93.60 & -1.61 \\
Oracle direction & 76.38 & 15.61 \\
Oracle angle & 86.29 & 5.70 \\
Joint oracle & 72.00 & 19.99 \\
\bottomrule
\end{tabular}
\end{table}

\subsection{Cross-Geometry Comparison on Matched 30-Task Sets}
\label{app:heldout-mesh}

To reduce reliance on semantic shortcuts and encourage visual-spatial reasoning, RoTA-Bench synthesizes 3D objects from 2D vector icons rather than drawing directly from common 3D CAD repositories, where familiar shapes may trigger pre-trained priors. To examine the effect of this design choice, we compare the original 30-task icon-derived ablation set with another 30-task set constructed from 30 distinct polygonal meshes from the Poly Haven CC0 asset collection. These meshes are based on real-world object categories and may therefore resemble objects encountered during model pretraining.

The two task sets have exactly matched reference-path length distributions. Both use the same point-splat renderer, auxiliary depth view, reversible 6--16-step initialization with $30^\circ$ rotations, 24-action space, full interaction history, 48-decision budget, success threshold, and termination rules. As shown in Table~\ref{tab:heldout-mesh-results}, GPT-5.6-Sol achieves a substantially higher mean final IoU on the Poly Haven mesh set and a slightly higher success rate than on the icon-derived set. In this matched-distribution comparison, tasks based on real-world object geometries yield higher observed performance than tasks involving geometries generated from 2D vector icons. Although the two sets are not paired task-by-task, the comparison suggests that the familiarity of geometric appearance may affect model performance.

\begin{table}[t]
\centering
\small
\setlength{\tabcolsep}{3pt}
\caption{GPT-5.6-Sol on two 30-task sets with exactly matched reference-path length distributions. The icon-derived set is the original ablation subset; the comparison set uses 30 distinct Poly Haven CC0 meshes based on real-world object categories.}
\label{tab:heldout-mesh-results}
\begin{tabular}{@{}lrrr@{}}
\toprule
Task set & Success & Mean final IoU & Mean steps \\
\midrule
Icon-derived tasks & 5/30 (16.7\%) & 0.5656 & 23.53 \\
Poly Haven meshes & 6/30 (20.0\%) & 0.7179 & 22.17 \\
\bottomrule
\end{tabular}
\end{table}

\section{Implementation and Evaluation Settings}
\label{app:implementation}
Table~\ref{tab:implementation} lists local software versions, hardware specifications, LLM hyperparameters, and evaluation settings.

\begin{table}[!t]
\centering\small
\caption{Software versions, hardware specifications, model hyperparameters, agent configurations, and interaction settings used for evaluation.}
\label{tab:implementation}
\begin{tabular}{p{0.32\linewidth}p{0.60\linewidth}}
\toprule
Component & Setting \\
\midrule
Rendering & Python 3.11.14; NumPy 2.3.3; OpenCV 4.13.0; PyTorch 2.10.0+cu130; CUDA 13.0 \\
Supporting packages & CairoSVG 2.8.2; cairocffi 1.7.1; Pillow 12.1.1; Numba 0.64.0 \\
API and agent packages & OpenAI 2.48.0; HTTPX 0.28.1; Microsoft Agent Framework core 1.12.1, OpenAI 1.11.0, orchestrations 1.0.1 \\
Local GPU server & 8 $\times$ NVIDIA A100-PCIE-40GB; 2 $\times$ Intel Xeon Silver 4314 (32 cores / 64 threads); 256 GB RAM; Driver 615.71.09 \\
Open-source model serving & SpatioLM-Understanding-InternVL3.5: \texttt{bfloat16}, greedy decoding (\texttt{do\_sample=False}), official architecture; local HTTP server \\
Response format & JSON; no streaming \\
LLM hyperparameters & Temperature 0; reasoning effort medium and seed 42 where supported; maximum 8,192 output tokens \\
Request timeout & 1,500 seconds \\
History and observations & Full interaction history; three $320\times320$ views in a $972\times320$ panel, separated by two 6-pixel gaps \\
Termination & Success at IoU $\geq0.9$, STOP, 48 decisions, or 12 consecutive rotations without a new best IoU \\
Agent tools & Python execution and image viewing; at most four tool calls in total per environment step \\
Magentic orchestration & One solver and one manager; at most four rounds and one reset \\
Request retries & Up to two outer retries; backoff parameter 1.6 seconds \\
Python tool limits & 300 seconds per call; 4 GiB memory; 64 MiB workspace; up to four returned images \\
Agent workspace & Task images and derived files; no internal poses or reference paths \\
\bottomrule
\end{tabular}
\end{table}

SpatioLM-Understanding-InternVL3.5 \citep{wu2026spatiolm} is evaluated using self-hosted local inference and incurs no commercial API token fees. Inference is hosted on an on-premise server equipped with NVIDIA A100-PCIE-40GB GPUs, dual Intel Xeon Silver 4314 processors (32 physical cores, 64 execution threads @ 2.40~GHz), and 256~GB of system RAM, running Ubuntu Linux with NVIDIA driver 615.71.09, CUDA 13.0, and PyTorch 2.10.0+cu130. The model is served through an internal OpenAI-compatible HTTP interface using the official \texttt{InternVL3RChatModel} architecture in \texttt{torch.bfloat16} precision. Generation uses deterministic greedy decoding (\texttt{do\_sample=False}, corresponding to temperature 0), matching the zero-temperature setting of closed-source models, with a context window upper limit of 8,192 output tokens.

Table~\ref{tab:api-costs} estimates API costs from recorded token usage and published list prices below, including Magentic's internal workflow and the three input ablation conditions. We use published list rates for GPT-6-Astra (\$10.00 / \$50.00 per million tokens) and GPT-5.6 (\$5.00 / \$30.00), base rates for Gemini and Qwen, and announced standard rates for Claude.\footnote{Official pricing sources: \url{https://developers.openai.com/api/docs/pricing/}, \url{https://ai.google.dev/gemini-api/docs/pricing}, \url{https://www.anthropic.com/news/claude-sonnet-5}, \url{https://www.anthropic.com/news/claude-opus-5}, and \url{https://www.alibabacloud.com/help/en/model-studio/model-pricing}.}

\begin{table}[!t]
\centering
\small
\setlength{\tabcolsep}{4.5pt}
\caption{Published API prices, recorded token counts, and estimated costs. Main evaluations use 100 tasks; the three GPT-5.6-Sol input ablations use the same 30-task subset. Rates are USD per million tokens using uncached input pricing; costs are in USD.}
\label{tab:api-costs}
\begin{tabular}{lrrrr}
\toprule
Condition &
\shortstack{Input/output\\rate} &
Input tokens &
Output tokens &
\shortstack{Total cost\\(USD)} \\
\midrule
GPT-5.6-Sol              & 5.00 / 30.00  & 28,883,151 &   853,753 & 170.0283 \\
GPT-5.6-Terra            & 2.50 / 15.00  & 27,905,743 & 1,062,892 &  85.7077 \\
GPT-5.6-Luna             & 1.00 / 6.00   & 22,630,532 & 1,087,401 &  29.1549 \\
GPT-6-Astra              & 10.00 / 50.00 & 29,975,049 & 1,187,433 & 359.1221 \\
Gemini 3.6 Flash         & 1.50 / 7.50   & 45,392,248 & 4,555,620 & 102.2555 \\
Claude Opus 5            & 5.00 / 25.00  & 17,014,514 &   700,389 & 102.5823 \\
Claude Sonnet 5          & 3.00 / 15.00  & 19,556,374 &   722,105 &  69.5007 \\
Qwen 3.7 Plus            & 0.40 / 1.60   & 10,719,193 &   901,152 &   5.7295 \\
GPT-5.6-Sol + ReAct      & 5.00 / 30.00  & 28,836,103 & 1,024,921 & 174.9281 \\
GPT-5.6-Sol + Magentic   & 5.00 / 30.00  & 49,032,175 & 2,208,867 & 311.4269 \\
\midrule
Sol: no aux. view        & 5.00 / 30.00  &  6,621,163 &   208,670 &  39.3659 \\
Sol: no history          & 5.00 / 30.00  &  2,065,970 &   305,183 &  19.4853 \\
Sol: neither input       & 5.00 / 30.00  &    691,672 &   310,963 &  12.7873 \\
\midrule
Total                    & ---           & 289,323,887 & 15,129,349 & 1,482.0747 \\
\bottomrule
\end{tabular}
\end{table}

\section{Prompt Examples}
\label{app:prompts}
\begin{figure}[!t]
\centering
\begin{minipage}{\linewidth}
\small
User input.
\begin{Verbatim}[breaklines=true,breakanywhere=true,breaksymbolleft={},fontsize=\footnotesize]
Current turn: 1/48.
Used steps: 0.
Remaining steps: 48.
The task ends automatically when the requirement is met.
If it continues, the current result has not yet met the requirement.
Current state to act on now:
\end{Verbatim}
\centerline{\includegraphics[width=0.56\linewidth]{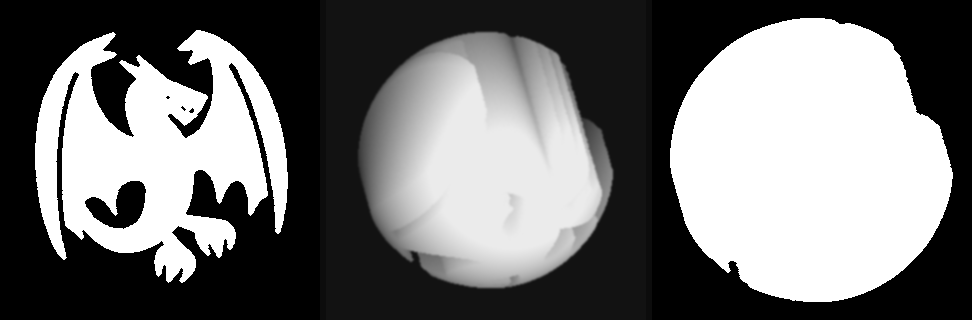}}
Example response.
\begin{Verbatim}[breaklines=true,breakanywhere=true,breaksymbolleft={},fontsize=\footnotesize]
{"cot":"...","type":"YAW_POS","angle":30}
\end{Verbatim}

\caption{First-turn prompt for single\_00627, showing the initial observation, user input, and an example JSON response.}
\label{fig:prompt-step1}
\end{minipage}
\end{figure}

\begin{figure}[!t]
\centering
\begin{minipage}{\linewidth}
\small
User input.
\begin{Verbatim}[breaklines=true,breakanywhere=true,breaksymbolleft={},fontsize=\footnotesize]
Current turn: 2/48.
Used steps: 1.
Remaining steps: 47.
The task ends automatically when the requirement is met.
If it continues, the current result has not yet met the requirement.
Historical state 1; action after this state:
step=1 action=YAW_POS angle=30 cot=...
\end{Verbatim}
\centerline{\includegraphics[width=0.50\linewidth]{figures/prompt_single_00627/panel_01.png}}
\begin{Verbatim}[breaklines=true,breakanywhere=true,breaksymbolleft={},fontsize=\footnotesize]
Current state to act on now:
\end{Verbatim}
\centerline{\includegraphics[width=0.50\linewidth]{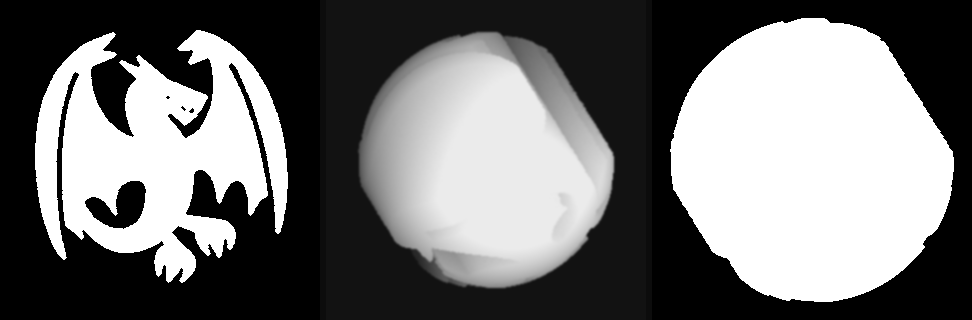}}
Example response.
\begin{Verbatim}[breaklines=true,breakanywhere=true,breaksymbolleft={},fontsize=\footnotesize]
{"cot":"...","type":"PITCH_NEG","angle":10}
\end{Verbatim}
\end{minipage}
\caption{Second-turn prompt for single\_00627, showing the first observation and action--COT record, current observation, user input, and an example JSON response.}
\label{fig:prompt-step2}
\end{figure}

\begin{figure}[!t]
\centering
\begin{minipage}{\linewidth}
\small
User input.
\begin{Verbatim}[breaklines=true,breakanywhere=true,breaksymbolleft={},fontsize=\footnotesize]
Current turn: 3/48.
Used steps: 2.
Remaining steps: 46.
The task ends automatically when the requirement is met.
If it continues, the current result has not yet met the requirement.
Historical state 1; action after this state:
step=1 action=YAW_POS angle=30 cot=...
\end{Verbatim}
\centerline{\includegraphics[width=0.48\linewidth]{figures/prompt_single_00627/panel_01.png}}
\begin{Verbatim}[breaklines=true,breakanywhere=true,breaksymbolleft={},fontsize=\footnotesize]
Historical state 2; action after this state:
step=2 action=PITCH_NEG angle=10 cot=...
\end{Verbatim}
\centerline{\includegraphics[width=0.48\linewidth]{figures/prompt_single_00627/panel_02.png}}
\begin{Verbatim}[breaklines=true,breakanywhere=true,breaksymbolleft={},fontsize=\footnotesize]
Current state to act on now:
\end{Verbatim}
\centerline{\includegraphics[width=0.48\linewidth]{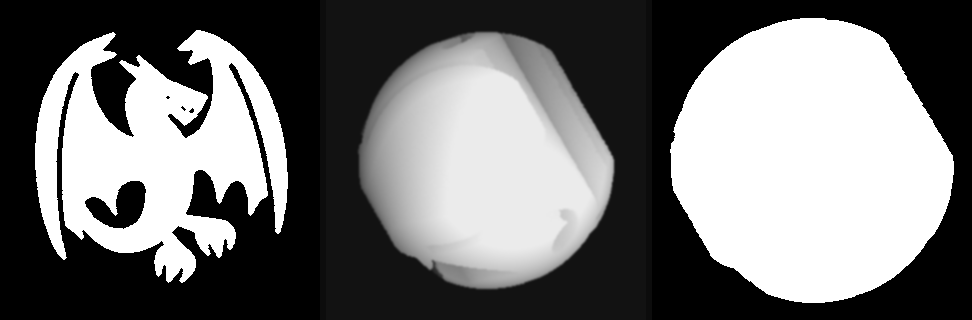}}
Example response.
\begin{Verbatim}[breaklines=true,breakanywhere=true,breaksymbolleft={},fontsize=\footnotesize]
{"cot":"...","type":"ROLL_POS","angle":5}
\end{Verbatim}
\end{minipage}
\caption{Third-turn prompt for single\_00627, showing both earlier observations and action--COT records, current observation, user input, and an example JSON response.}
\label{fig:prompt-step3}
\end{figure}

The following is a prompt example for single\_00627 (Figures~\ref{fig:prompt-step1}--\ref{fig:prompt-step3}). We call the depth-enhanced auxiliary observation the ``oblique view'' because it exposes depth and occlusion cues that are absent from the binary projection, similar in purpose to an oblique depiction of a 3D shape. The name describes its visual role rather than a camera rotation: the view is rendered with the same camera direction and object orientation as the current projection, while depth-dependent shading provides the additional cues. Line breaks below are for readability.

System prompt.
{\small
\begin{quote}
You are playing a deterministic turn-based 3D-to-2D alignment game. Rotate only the CURRENT object so its projection matches TARGET within the step budget. Each observation provides images and the current task feedback. Each panel is one image ordered left-to-right as TARGET projection | CURRENT oblique view | CURRENT projection. The oblique view is an auxiliary 3D pose cue; compare TARGET with CURRENT projection for alignment. Rotations use observer/camera coordinates: PITCH\_POS=pitch upward, PITCH\_NEG=pitch downward, YAW\_POS=turn toward observer right, YAW\_NEG=turn toward observer left, ROLL\_POS=clockwise in the image plane, ROLL\_NEG=counter-clockwise. Allowed action tokens: PITCH\_POS, PITCH\_NEG, YAW\_POS, YAW\_NEG, ROLL\_POS, ROLL\_NEG, STOP. Allowed angles: 30, 10, 5, 1. History is chronological. Each historical panel is the state before the action written immediately beside it. The environment ends automatically when the requirement is met. Use STOP only when no actionable improvement idea remains; do not use STOP merely because the silhouettes appear aligned. Return exactly one JSON object and no extra text: \{\ttfamily "cot":"brief current-step reasoning","type":"TOKEN","angle":ANGLE\}. The type and angle must come from the allowed sets.
\end{quote}
}

\section{Success and Failure Examples}
\label{app:cases}
We select the GPT-5.6-Sol run with the fewest executed rotations separately among successful and unsuccessful tasks, breaking ties by ascending task identifier. The selected runs are single\_00270 (8 rotations) and single\_00039 (12 rotations). These are short illustrative cases rather than representative samples of all model behavior.

In the successful case (Figure~\ref{fig:case-success}), the model reverses a yaw after it reduces alignment, then changes the roll and makes smaller pitch and yaw adjustments. IoU rises from 0.624 to 0.977. The sequence illustrates recovery from an unhelpful action followed by refinement across axes. In the failed case (Figure~\ref{fig:case-failure}), all 12 actions are $30^\circ$ rotations in the same yaw direction, completing a full turn and returning to the initial alignment (IoU 0.450). The model repeatedly proposes revealing a hidden face without changing its action strategy; no rotation improves on the initial IoU, so stagnation stopping ends the run.
\begin{figure}[!t]
\centering
\begin{minipage}[t]{0.485\linewidth}
\vspace{0pt}\footnotesize
Rotation 1: IoU $0.624\rightarrow0.647$.\par
\includegraphics[width=\linewidth]{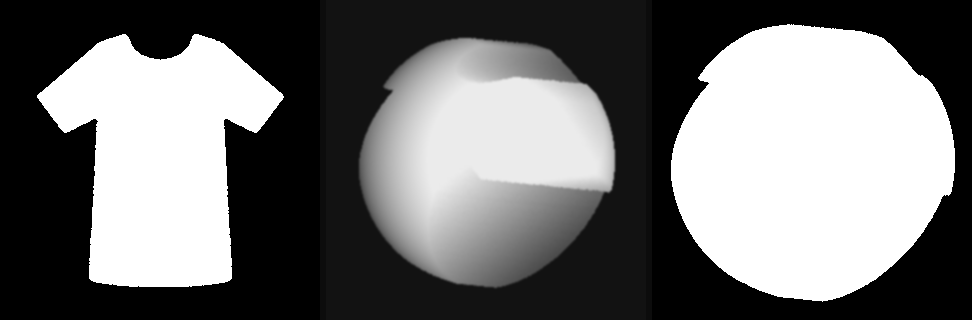}
\begin{Verbatim}[breaklines=true,breakanywhere=true,breaksymbolleft={},fontsize=\footnotesize]
{
  "cot": "The target is a broad
front-facing shirt silhouette, while the
current projection is nearly circular.
The oblique cue suggests the garment is
turned substantially away from the
target, so begin with a moderate yaw
toward the observer's right.",
  "type": "YAW_POS",
  "angle": 30
}
\end{Verbatim}
\end{minipage}
\hfill
\begin{minipage}[t]{0.485\linewidth}
\vspace{0pt}\footnotesize
Rotation 2: IoU $0.647\rightarrow0.651$.\par
\includegraphics[width=\linewidth]{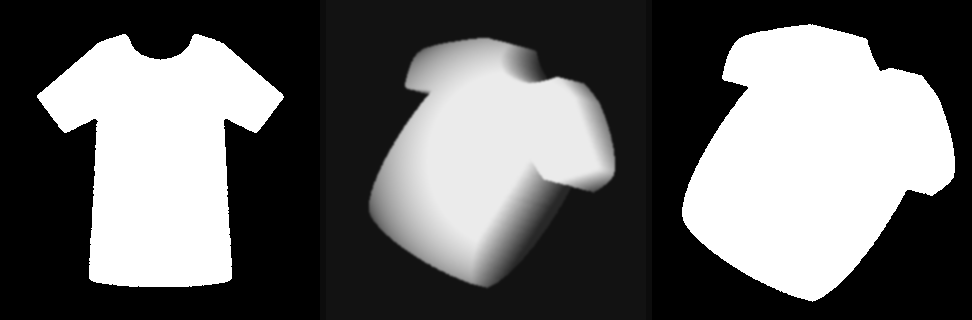}
\begin{Verbatim}[breaklines=true,breakanywhere=true,breaksymbolleft={},fontsize=\footnotesize]
{
  "cot": "The first yaw improved the
silhouette from circular to a
recognizable garment, but it remains
strongly oblique and narrow. Continue
yawing toward the observer's right to
bring the front projection closer to the
broad target.",
  "type": "YAW_POS",
  "angle": 30
}
\end{Verbatim}
\end{minipage}
\par\vspace{6pt}
\begin{minipage}[t]{0.485\linewidth}
\vspace{0pt}\footnotesize
Rotation 3: IoU $0.651\rightarrow0.633$.\par
\includegraphics[width=\linewidth]{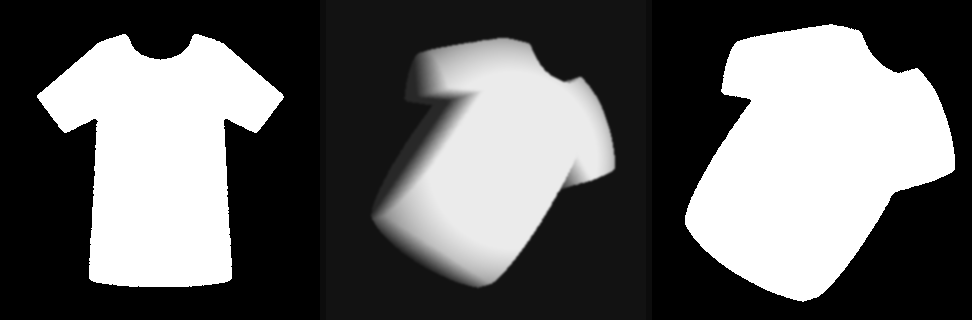}
\begin{Verbatim}[breaklines=true,breakanywhere=true,breaksymbolleft={},fontsize=\footnotesize]
{
  "cot": "The projection is improving
with yaw but remains strongly oblique
and diagonally oriented, so continue
turning toward the observer's right.",
  "type": "YAW_POS",
  "angle": 30
}
\end{Verbatim}
\end{minipage}
\hfill
\begin{minipage}[t]{0.485\linewidth}
\vspace{0pt}\footnotesize
Rotation 4: IoU $0.633\rightarrow0.651$.\par
\includegraphics[width=\linewidth]{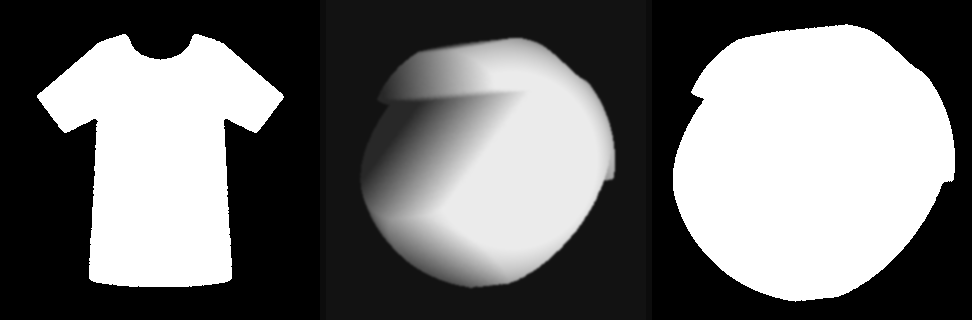}
\begin{Verbatim}[breaklines=true,breakanywhere=true,breaksymbolleft={},fontsize=\footnotesize]
{
  "cot": "The projection has become
circular again after continued yaw,
suggesting the useful yaw range was
passed. Reverse yaw to recover a broader
garment silhouette.",
  "type": "YAW_NEG",
  "angle": 30
}
\end{Verbatim}
\end{minipage}
\caption{GPT-5.6-Sol's successful trajectory on single\_00270: rotation steps 1--4 of 8, with pre-action observations and original JSON responses. Panels show the target projection, auxiliary view, and current projection from left to right; IoU annotations were not provided to the model.}
\label{fig:case-success}
\end{figure}

\begin{figure}[!t]
\centering
\ContinuedFloat
\begin{minipage}[t]{0.485\linewidth}
\vspace{0pt}\footnotesize
Rotation 5: IoU $0.651\rightarrow0.792$.\par
\includegraphics[width=\linewidth]{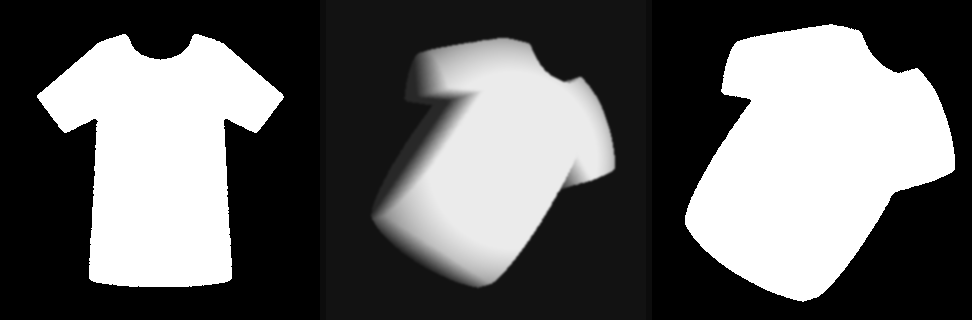}
\begin{Verbatim}[breaklines=true,breakanywhere=true,breaksymbolleft={},fontsize=\footnotesize]
{
  "cot": "The garment is still
diagonally oriented, with its long axis
slanting down-left instead of vertically
like the target. Apply a moderate
counter-clockwise roll to correct the
image-plane orientation.",
  "type": "ROLL_NEG",
  "angle": 30
}
\end{Verbatim}
\end{minipage}
\hfill
\begin{minipage}[t]{0.485\linewidth}
\vspace{0pt}\footnotesize
Rotation 6: IoU $0.792\rightarrow0.823$.\par
\includegraphics[width=\linewidth]{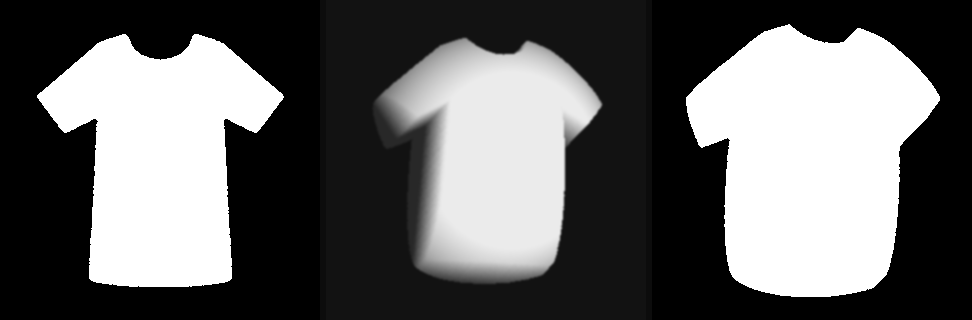}
\begin{Verbatim}[breaklines=true,breakanywhere=true,breaksymbolleft={},fontsize=\footnotesize]
{
  "cot": "The projection is nearly
front-facing and correctly rolled. Its
vertical extent is slightly too tall,
with the shoulders too high and the
torso bottom too low, so apply a small
downward pitch to compress the vertical
silhouette toward the target.",
  "type": "PITCH_NEG",
  "angle": 10
}
\end{Verbatim}
\end{minipage}
\par\vspace{6pt}
\begin{minipage}[t]{0.485\linewidth}
\vspace{0pt}\footnotesize
Rotation 7: IoU $0.823\rightarrow0.892$.\par
\includegraphics[width=\linewidth]{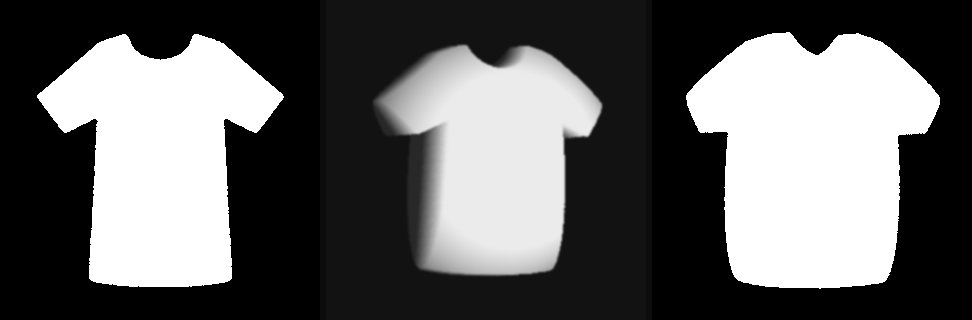}
\begin{Verbatim}[breaklines=true,breakanywhere=true,breaksymbolleft={},fontsize=\footnotesize]
{
  "cot": "The current silhouette is very
close, but the torso is slightly too
wide and the sleeve undersides sit lower
than the target. A small yaw adjustment
is the most actionable refinement while
preserving the nearly correct vertical
scale.",
  "type": "YAW_NEG",
  "angle": 5
}
\end{Verbatim}
\end{minipage}
\hfill
\begin{minipage}[t]{0.485\linewidth}
\vspace{0pt}\footnotesize
Rotation 8: IoU $0.892\rightarrow0.977$.\par
\includegraphics[width=\linewidth]{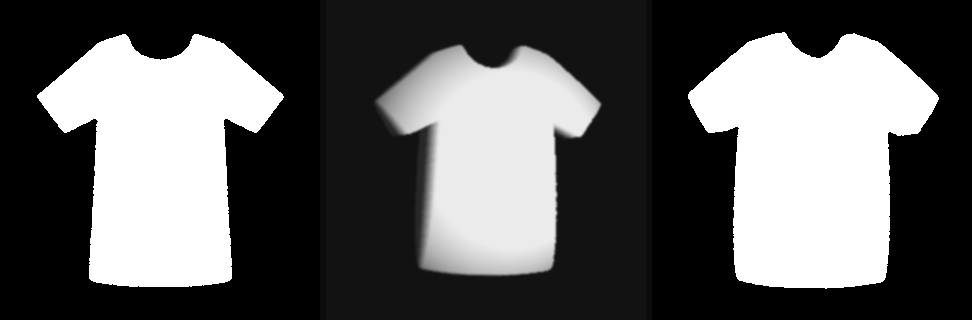}
\begin{Verbatim}[breaklines=true,breakanywhere=true,breaksymbolleft={},fontsize=\footnotesize]
{
  "cot": "The projection is very close,
but the current torso remains slightly
wider than the target and the
sleeve-to-torso transitions sit a little
low. Continue the same small yaw
adjustment to narrow the torso while
preserving the overall alignment.",
  "type": "YAW_NEG",
  "angle": 5
}
\end{Verbatim}
\end{minipage}
\par\vspace{4pt}
{\footnotesize Final state after rotation 8: IoU 0.977.\par}
\includegraphics[width=0.485\linewidth]{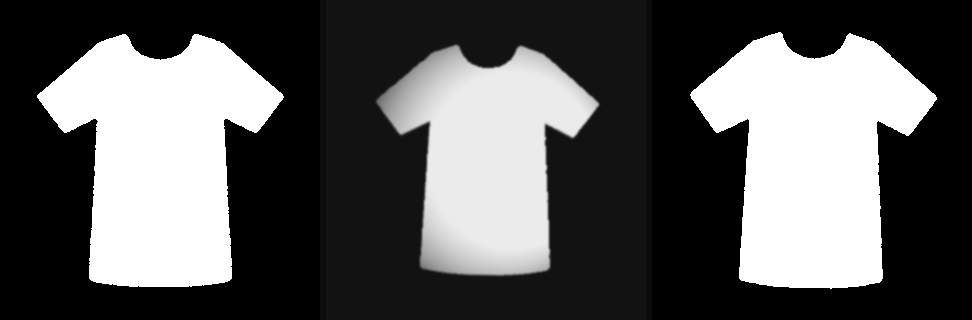}
\caption{GPT-5.6-Sol's successful trajectory on single\_00270: rotation steps 5--8 of 8 and the final state, with pre-action observations and original JSON responses. Panels show the target projection, auxiliary view, and current projection from left to right; IoU annotations were not provided to the model.}
\label{fig:case-success-2}
\end{figure}

\begin{figure}[!t]
\centering
\begin{minipage}[t]{0.485\linewidth}
\vspace{0pt}\footnotesize
Rotation 1: IoU $0.450\rightarrow0.441$.\par
\includegraphics[width=\linewidth]{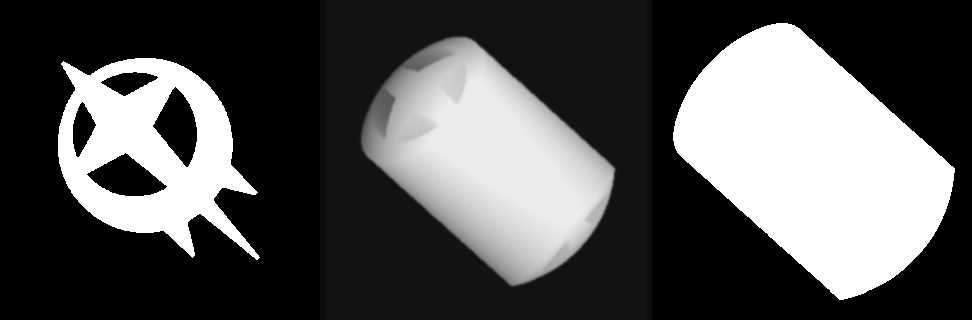}
\begin{Verbatim}[breaklines=true,breakanywhere=true,breaksymbolleft={},fontsize=\footnotesize]
{
  "cot": "The target is an end-on view
while the current projection is a side
view. The object's axis projects
down-right, so yawing left should reduce
its rightward projection.",
  "type": "YAW_NEG",
  "angle": 30
}
\end{Verbatim}
\end{minipage}
\hfill
\begin{minipage}[t]{0.485\linewidth}
\vspace{0pt}\footnotesize
Rotation 2: IoU $0.441\rightarrow0.427$.\par
\includegraphics[width=\linewidth]{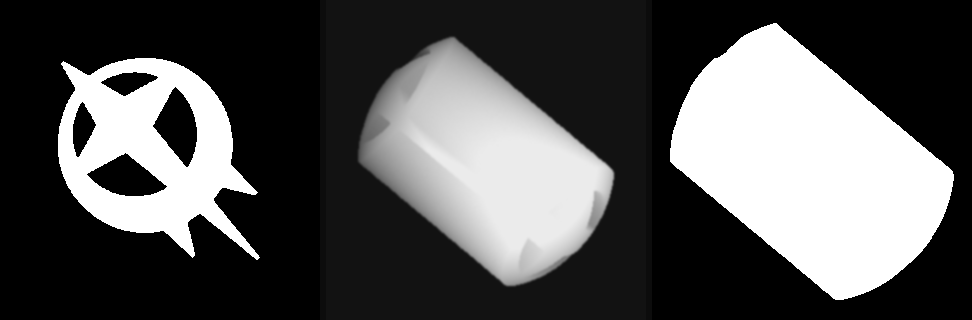}
\begin{Verbatim}[breaklines=true,breakanywhere=true,breaksymbolleft={},fontsize=\footnotesize]
{
  "cot": "The current projection is
still a side view with the circular face
hidden. Continue yawing left to bring
the object's axis toward the viewer and
expose the detailed end face.",
  "type": "YAW_NEG",
  "angle": 30
}
\end{Verbatim}
\end{minipage}
\par\vspace{6pt}
\begin{minipage}[t]{0.485\linewidth}
\vspace{0pt}\footnotesize
Rotation 3: IoU $0.427\rightarrow0.377$.\par
\includegraphics[width=\linewidth]{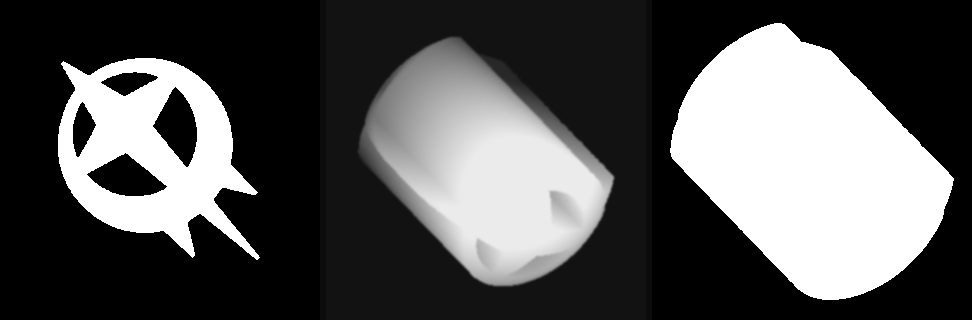}
\begin{Verbatim}[breaklines=true,breakanywhere=true,breaksymbolleft={},fontsize=\footnotesize]
{
  "cot": "The current projection is
still a side-on cylinder with the
detailed end face hidden. Continue
yawing left to bring the axis toward the
viewer and reveal the target end-face
pattern.",
  "type": "YAW_NEG",
  "angle": 30
}
\end{Verbatim}
\end{minipage}
\hfill
\begin{minipage}[t]{0.485\linewidth}
\vspace{0pt}\footnotesize
Rotation 4: IoU $0.377\rightarrow0.378$.\par
\includegraphics[width=\linewidth]{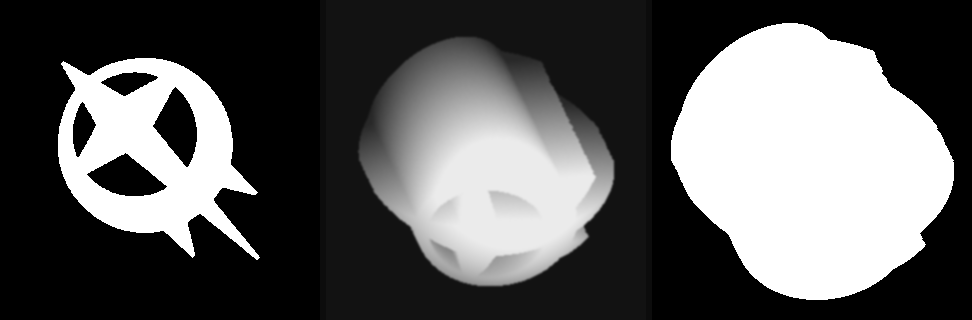}
\begin{Verbatim}[breaklines=true,breakanywhere=true,breaksymbolleft={},fontsize=\footnotesize]
{
  "cot": "The projection is now nearly
end-on, but the target's internal
end-face pattern is still not visible.
Continue yawing left to bring the
patterned face fully toward the
viewer.",
  "type": "YAW_NEG",
  "angle": 30
}
\end{Verbatim}
\end{minipage}
\caption{GPT-5.6-Sol's unsuccessful trajectory on single\_00039: rotation steps 1--4 of 12, with pre-action observations and original JSON responses. Panels show the target projection, auxiliary view, and current projection from left to right; IoU annotations were not provided to the model.}
\label{fig:case-failure}
\end{figure}

\begin{figure}[!t]
\centering
\ContinuedFloat
\begin{minipage}[t]{0.485\linewidth}
\vspace{0pt}\footnotesize
Rotation 5: IoU $0.378\rightarrow0.395$.\par
\includegraphics[width=\linewidth]{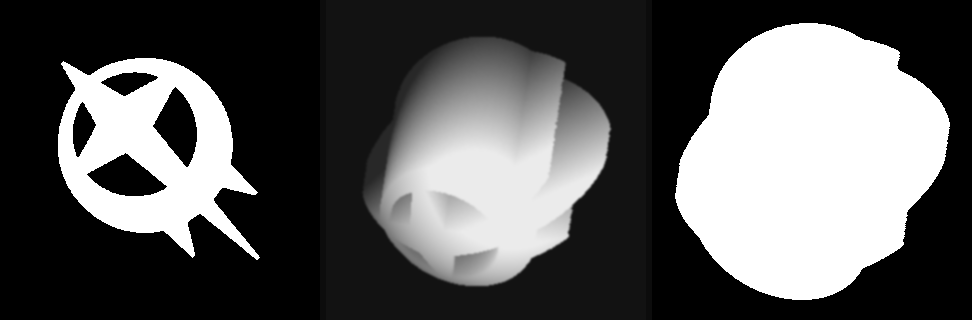}
\begin{Verbatim}[breaklines=true,breakanywhere=true,breaksymbolleft={},fontsize=\footnotesize]
{
  "cot": "The projection is nearly
face-on but the patterned end face is
still obscured. Continue yawing left to
expose the target pattern.",
  "type": "YAW_NEG",
  "angle": 30
}
\end{Verbatim}
\end{minipage}
\hfill
\begin{minipage}[t]{0.485\linewidth}
\vspace{0pt}\footnotesize
Rotation 6: IoU $0.395\rightarrow0.380$.\par
\includegraphics[width=\linewidth]{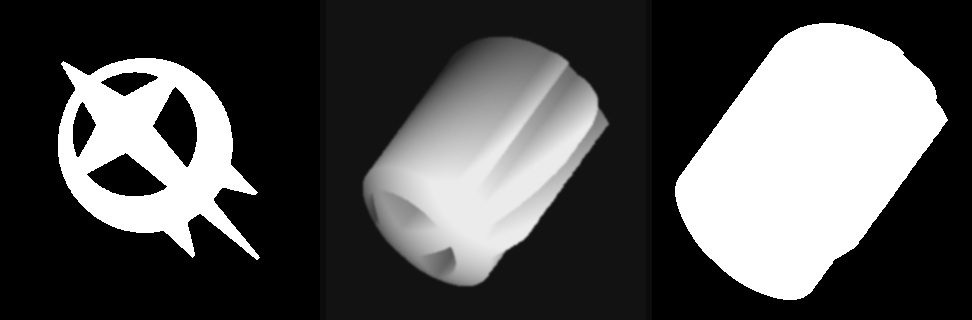}
\begin{Verbatim}[breaklines=true,breakanywhere=true,breaksymbolleft={},fontsize=\footnotesize]
{
  "cot": "The patterned end face remains
hidden in the current projection.
Continue yawing left to bring that face
toward the viewer.",
  "type": "YAW_NEG",
  "angle": 30
}
\end{Verbatim}
\end{minipage}
\par\vspace{6pt}
\begin{minipage}[t]{0.485\linewidth}
\vspace{0pt}\footnotesize
Rotation 7: IoU $0.380\rightarrow0.371$.\par
\includegraphics[width=\linewidth]{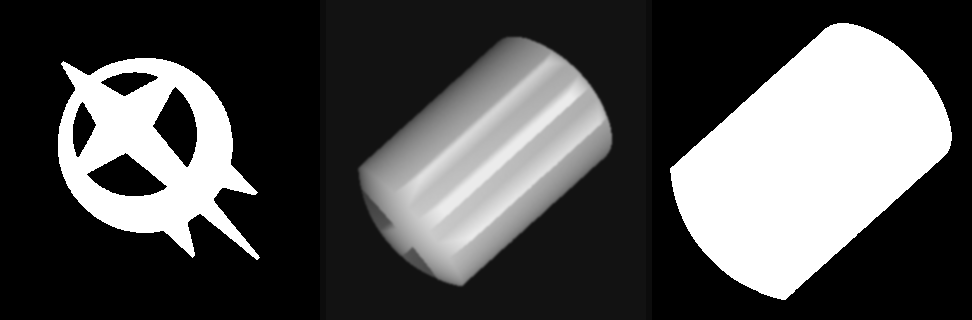}
\begin{Verbatim}[breaklines=true,breakanywhere=true,breaksymbolleft={},fontsize=\footnotesize]
{
  "cot": "The patterned face is still
not visible and the current projection
remains side-on. Continue yawing left to
rotate the patterned end toward the
viewer.",
  "type": "YAW_NEG",
  "angle": 30
}
\end{Verbatim}
\end{minipage}
\hfill
\begin{minipage}[t]{0.485\linewidth}
\vspace{0pt}\footnotesize
Rotation 8: IoU $0.371\rightarrow0.369$.\par
\includegraphics[width=\linewidth]{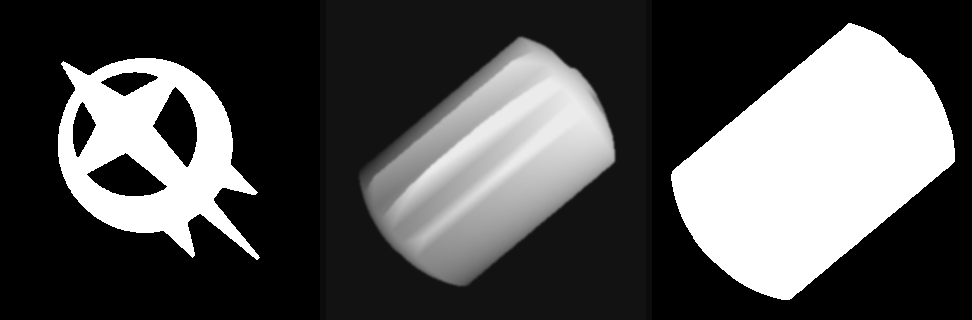}
\begin{Verbatim}[breaklines=true,breakanywhere=true,breaksymbolleft={},fontsize=\footnotesize]
{
  "cot": "The patterned end face is
still not visible and the current
projection remains side-on. Continue
yawing left to rotate the patterned face
toward the viewer.",
  "type": "YAW_NEG",
  "angle": 30
}
\end{Verbatim}
\end{minipage}
\caption{GPT-5.6-Sol's unsuccessful trajectory on single\_00039: rotation steps 5--8 of 12, with pre-action observations and original JSON responses. Panels show the target projection, auxiliary view, and current projection from left to right; IoU annotations were not provided to the model.}
\label{fig:case-failure-2}
\end{figure}

\begin{figure}[!t]
\centering
\ContinuedFloat
\begin{minipage}[t]{0.485\linewidth}
\vspace{0pt}\footnotesize
Rotation 9: IoU $0.369\rightarrow0.364$.\par
\includegraphics[width=\linewidth]{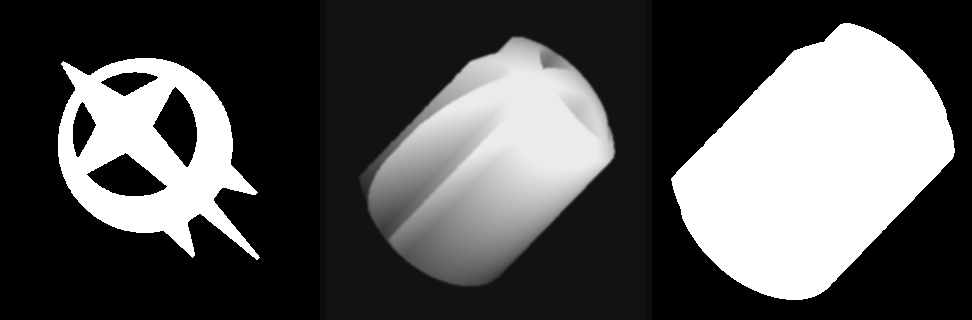}
\begin{Verbatim}[breaklines=true,breakanywhere=true,breaksymbolleft={},fontsize=\footnotesize]
{
  "cot": "The current projection is
still side-on and lacks the target's
patterned end face. Continue yawing left
to bring the end face toward the
viewer.",
  "type": "YAW_NEG",
  "angle": 30
}
\end{Verbatim}
\end{minipage}
\hfill
\begin{minipage}[t]{0.485\linewidth}
\vspace{0pt}\footnotesize
Rotation 10: IoU $0.364\rightarrow0.379$.\par
\includegraphics[width=\linewidth]{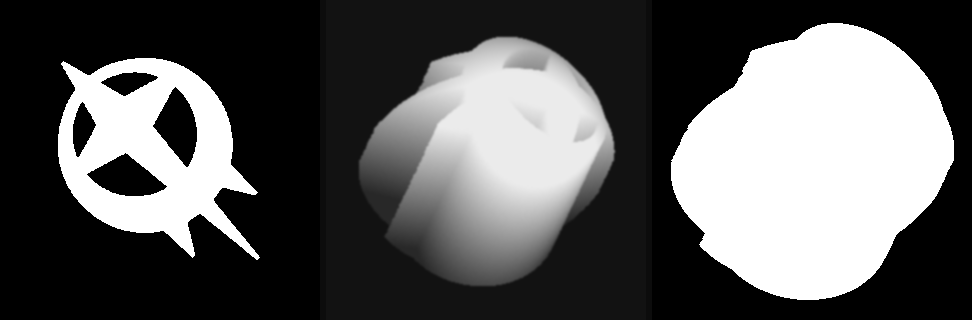}
\begin{Verbatim}[breaklines=true,breakanywhere=true,breaksymbolleft={},fontsize=\footnotesize]
{
  "cot": "The patterned end face is
beginning to face the viewer, but the
current projection still lacks the
target openings. Continue yawing left to
expose and align the patterned face.",
  "type": "YAW_NEG",
  "angle": 30
}
\end{Verbatim}
\end{minipage}
\par\vspace{6pt}
\begin{minipage}[t]{0.485\linewidth}
\vspace{0pt}\footnotesize
Rotation 11: IoU $0.379\rightarrow0.437$.\par
\includegraphics[width=\linewidth]{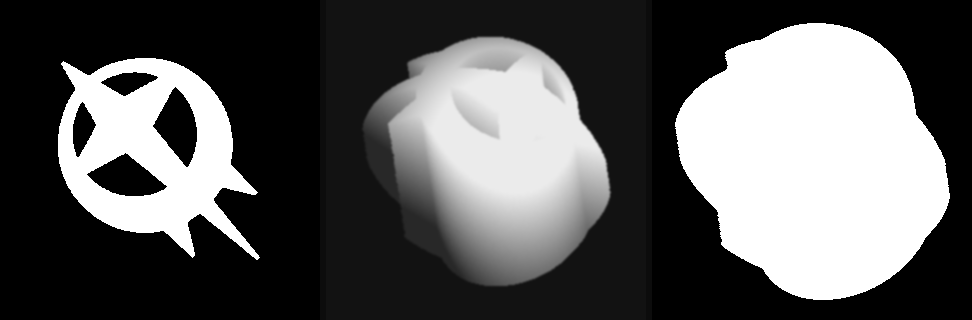}
\begin{Verbatim}[breaklines=true,breakanywhere=true,breaksymbolleft={},fontsize=\footnotesize]
{
  "cot": "The current silhouette is
close to the target, and the patterned
end face is visible obliquely but not
yet in the projection. Continue yawing
left to bring that face fully toward the
viewer.",
  "type": "YAW_NEG",
  "angle": 30
}
\end{Verbatim}
\end{minipage}
\hfill
\begin{minipage}[t]{0.485\linewidth}
\vspace{0pt}\footnotesize
Rotation 12: IoU $0.437\rightarrow0.450$.\par
\includegraphics[width=\linewidth]{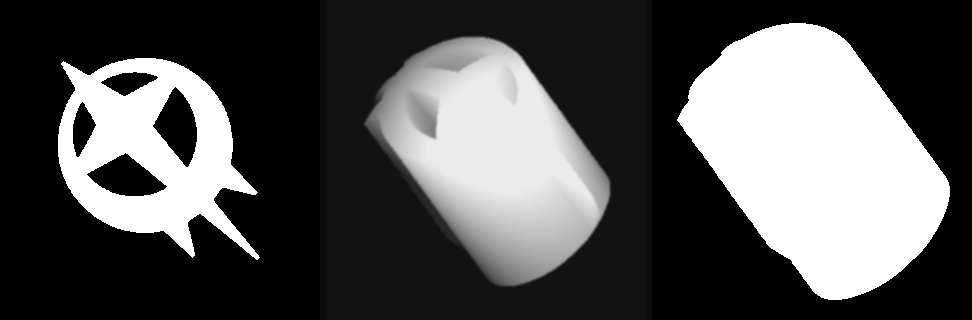}
\begin{Verbatim}[breaklines=true,breakanywhere=true,breaksymbolleft={},fontsize=\footnotesize]
{
  "cot": "The patterned face is still
not visible in the current projection,
so continue yawing left to bring it
fully toward the viewer.",
  "type": "YAW_NEG",
  "angle": 30
}
\end{Verbatim}
\end{minipage}
\par\vspace{4pt}
{\footnotesize Final state after rotation 12: IoU 0.450.\par}
\includegraphics[width=0.485\linewidth]{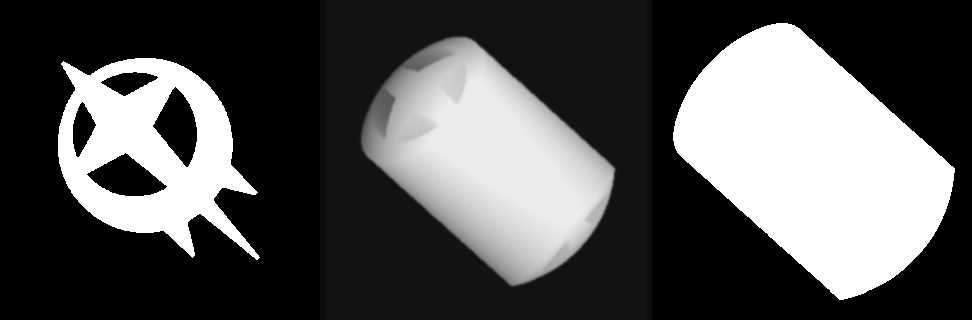}
\caption{GPT-5.6-Sol's unsuccessful trajectory on single\_00039: rotation steps 9--12 of 12 and the final state, with pre-action observations and original JSON responses. Panels show the target projection, auxiliary view, and current projection from left to right; IoU annotations were not provided to the model.}
\label{fig:case-failure-3}
\end{figure}
\end{document}